\documentclass[journal]{IEEEtran}

\usepackage[pdftex]{graphicx}
\usepackage{graphicx}

\usepackage{algorithmicx}
\usepackage{algpseudocode}
\usepackage{algorithm}
\usepackage{amssymb}
\usepackage{cite}
\usepackage{color}
\usepackage[cmex10]{amsmath}

\usepackage{balance}
\usepackage{url}
\usepackage{enumerate}
\usepackage{hyperref}

\usepackage{eqparbox}

\algnewcommand\algorithmicswitch{\textbf{switch}}
\algnewcommand\algorithmiccase{\textbf{case}}
\algnewcommand\algorithmicassert{\texttt{assert}}
\algnewcommand\Assert[1]{\State \algorithmicassert(#1)}

\algdef{SE}[SWITCH]{Switch}{EndSwitch}[1]{\algorithmicswitch\ #1\ \algorithmicdo}{\algorithmicend\ \algorithmicswitch}
\algdef{SE}[CASE]{Case}{EndCase}[1]{\algorithmiccase\ #1}{\algorithmicend\ \algorithmiccase}
\algtext*{EndSwitch}
\algtext*{EndCase}

\newdimen{\algindent}
\algnewcommand\LeftComment[2]{
\hspace{#1\algindent}$\triangleright$ \eqparbox{COMMENT}{#2} \hfill 
}

\newtheorem{theorem}{\textbf{Theorem}}[section]
\newtheorem{definition}{\textbf{Definition}}[section]
\newtheorem{lemma}{\textbf{Lemma}}[section]
\newtheorem{proposition}{\textbf{Proposition}}[section]

\newtheorem{assumption}{\textbf{Assumption}}[section]
\newtheorem{problem}{\textbf{Problem}}[section]
\newtheorem{remark}{\textbf{Remark}}[section]
\newtheorem{algo}{\textbf{Algorithm}}[section]

\newcommand{\oprocendsymbol}{\hbox{$\bullet$}}
\newcommand{\oprocend}{\relax\ifmmode\else\unskip\hfill\fi\oprocendsymbol}

\title{\LARGE \bf
Intermittent Connectivity Maintenance with  Heterogeneous Robots}

\author{Rosario Aragues$^{1,3}$, Dimos V. Dimarogonas$^{2}$, Pablo Guallar$^{1}$ and Carlos Sagues$^{1}$
\thanks{*Supported by COMMANDIA SOE2/P1/F0638 (Interreg Sudoe Programme, ERDF), CAS18/00082, Ministerio de Ciencia,  Innovaci{\'o}n y Universidades,
PGC2018-098719-B-I00 (MCIU/AEI/FEDER, UE), DGA T45-17R (Gobierno de Arag{\'o}n), and the Knut och Alice Wallenberg Foundation (KAW)
}
\thanks{$^{1}$R. Aragues, P. Guallar and C. Sagues are with DIIS Universidad de Zaragoza and 
Instituto de Investigaci{\'o}n en Ingenier{\'i}a de Arag{\'o}n I3A, Spain
        {\tt\small raragues@unizar.es}, {\tt\small pabloguallar@gmail.com}, {\tt\small csagues@unizar.es}}
\thanks{$^{2}$D. V. Dimarogonas is with the School of Electrical Engineering and Computer Science, KTH, Stockholm, Sweden {\tt\small dimos@kth.se}}
  \thanks{\textcolor{red}{This is the accepted version of the manuscript: R. Aragues, D. V. Dimarogonas, P. Guallar and C. Sagues, ``Intermittent Connectivity Maintenance With Heterogeneous Robots," in IEEE Transactions on Robotics, vol. 37, no. 1, pp. 225-245, Feb. 2021, doi: 10.1109/TRO.2020.3014521.
\textbf{Please cite the publisher's version}. For the publisher's version and full citation details see:\\
\protect\url{https://doi.org/10.1109/TRO.2020.3014521}. 
}}
 \thanks{© 2021 IEEE.  Personal use of this material is permitted.  Permission from IEEE must be obtained for all other uses, in any current or future media, including reprinting/republishing this material for advertising or promotional purposes, creating new collective works, for resale or redistribution to servers or lists, or reuse of any copyrighted component of this work in other works.}
\thanks{$^{3}$Corresponding author.}
}

\begin{document}

\maketitle

\begin{abstract}
We consider a scenario of cooperative task servicing, with a team of 
heterogeneous robots with different maximum speeds and communication radii, in charge of keeping the network intermittently connected. 
We abstract the task locations into a $1D$ cycle graph that is traversed by the communicating robots, and we discuss intermittent communication 
 strategies so that each task location is periodically visited, with a worst--case revisiting time. 
Robots move forward and backward along the cycle graph, exchanging data with their previous and next neighbors when they meet, and updating their region boundaries. Asymptotically, each robot is in charge of a  region of the cycle graph, depending on its  capabilities. The method is distributed, and robots only exchange data when they meet.
\end{abstract}


\IEEEpeerreviewmaketitle

\section{Introduction}

Servicing tasks is a core multi--robot application \cite{Guo-TAC2016}. 
We consider a cooperative task servicing scenario, with a team of task--robots, 
visiting different task locations to service them, and a team of communicating--robots  
in charge of keeping the task locations intermittently connected. Considering heterogeneous teams of robots with different roles and aims has a long history, e.g., \cite{guo2017distributed}. Here, when 
a task--robot wants to propagate and get data updates, it waits at the current task  place for a communicating--robot to show up, it exchanges data, and then moves to the next task location. 
The problem of visiting tasks located on the plane can be abstracted into a  $1D$ scenario by building a cycle graph  connecting the task locations \cite{gabriely2001spanning,applegate2006traveling}. We focus on the  coordination of the communicating--robots on this cycle graph, which are heterogeneous and have different maximum speeds and communication radii. 

The problem of connectivity control has received a lot of attention during the last years. A review of several methods can be found at \cite{Zavlanos-ProcIEEE2011}. A first approach consists of keeping the network connected at all times. This can be achieved by keeping the initial set of links, with possible link additions \cite{Boskos-JCO2017,Guo-TAC2016}, or by keeping pairs of links according to some underlying topology which is updated  as robots move, for undirected \cite{Soleymani-15ACC,Schuresko-JCO2012,Aranda-16ACC}, or directed graphs \cite{Poonawala-tac2017}. Several works use global connectivity approaches \cite{Gasparri-TRO2017,Nestmeyer-AuRo2017}, that rely on global parameters like the algebraic connectivity and Fiedler eigenvector. These works usually encode additional terms on the model, like obstacle of inter--robot collision avoidance, and they often study the performance degradation of the high--level task due to the effect of the  connectivity maintenance action. Depending on the environment size, and the  amount of robots and their capabilities, it may not be possible to accomplish a high--level task using a strategy based on keeping the network connected at all times.

An alternative are intermittent connectivity scenarios \cite{Kantaros-TAC2017,kantaros2019temporal,KhodayiKantarosTRO2019}. The network may be disconnected at every time instant, but it is jointly connected \emph{over time} and infinitely often. The key idea is to design the robot motions to ensure this behavior. These methods are more flexible, since they are always guaranteed to work, even if the environment size is larger, at the cost of  performance degradation. One of the  notable approaches on intermittent  connectivity is \cite{Kantaros-TAC2017}, where the goal is to ensure the connectivity on an environmental graph, by making robots move forward and backward on the links of this graph. For the method to work properly, the number of robots must equal the number of links in the graph. In addition, since each robot is trapped on its associated edge, the method cannot 
take any advantage from heterogeneous robots with larger maximum speeds and communication radii, which have to move slower depending on the worst--case robot motion (the slowest robot and / or the one assigned to the largest link). 

In \cite{kantaros2019temporal,KhodayiKantarosTRO2019} robots are not restricted to the links of a fixed graph. Robots are organized in teams, 
and they meet at a point in the environment chosen by the team members  \cite{kantaros2019temporal}, or form connected sub-networks in the space \cite{KhodayiKantarosTRO2019} to exchange data. 
Some robots belong to more than one team, and the team graph must be connected. Although there is more flexibility,  \cite{kantaros2019temporal,KhodayiKantarosTRO2019} require the robot teams to be selected by the user and they also require the offline schedule of the communication events. Thus,  \cite{kantaros2019temporal,KhodayiKantarosTRO2019} do  not take  advantage of the improved capabilities of individual robots. Moreover, 
 \cite{Kantaros-TAC2017,kantaros2019temporal,KhodayiKantarosTRO2019} do not 
self adapt to communicating robots entering and leaving the communicating team or varying their communication radii and maximum speeds. 
In \cite{GuoTRO2018} another example of application of intermittent connectivity ideas is presented. There, there are robots in charge of gathering data, 
 with limited buffer capabilities, that meet with some relay robots to upload the data. 
Compared to our work, in \cite{GuoTRO2018} the cooperation between relay robots in charge of the communication is weaker, since it only happens due to spontaneous (unplanned) meetings.

We propose an intermittent connectivity strategy where the robots move forward and  backward on the $1D$ cycle graph of the environment. Each robot has two neighbors in the cycle graph, and robots exchange data when they meet at the boundaries of their assigned regions. Robots which are faster or with larger communicating radius, are in charge of larger regions in the cycle graph. These regions are updated online in a distributed way, using local data on the involved robots.

The proposed method is similar to a \emph{beads--on--a--ring} strategy 
\cite{wang2010synchronization,susca2007synchronization,susca2014synchronization}, where each robot moves forward and backward on a specific region of the ring, impacting with its previous and next neighbors, and exchanging data only during the impacts. However, in beads--on--a--ring methods, the aim is that the robots synchronize to move at the same speed, which may be pre--established \cite{wang2010synchronization}, or may depend on the average of the initial robot speeds \cite{susca2007synchronization}, \cite{susca2014synchronization}, and end up covering regions of equal length. Here instead, the aim is that robots are in charge of larger regions if they are faster or have larger communication radii. 
In addition, \cite{wang2010synchronization,susca2007synchronization,susca2014synchronization} let robots to speed up without restrictions.  In the proposed method  robots cannot move faster than their maximum speed, so the strategy and approach differs to accommodate for this restriction. 
Our work is also related to works on coverage over a ring~\cite{song2016coverage}, although there the aim is to make robots converge to fixed points with associated coverage regions, instead of making them move forward and backward. The assumptions on the data exchange and on the communication capabilities are different in both scenarios, and so are the methods used.

The contributions of this paper are: 
$(i)$ a distributed method that does not depend on a specific number of robots, that takes advantage of the heterogeneous nature of the robots, and that only requires data exchange during robot meetings; 
$(ii)$ the proof that, asymptotically, each robot is in charge of a region with a length depending on its maximum speed and communication radius; 
$(iii)$ the proof of convergence to configurations with performance guaranties; and 
$(iv)$ the validation of the method in a realistic simulation environment using ROS/Gazebo.

A preliminary version of this work appears in \cite{Aragues-19ACC}. Compared to \cite{Aragues-19ACC}, here we make a thorough study of the performance of the method in terms of the revisiting times of the locations on the cycle graph (Theorems~\ref{th_performance_balanced} and \ref{th_performance_unbalanced}). In order to prove these theorems, we build on several theoretical results, that are developed along Sections  \ref{sec_interlaced} and \ref{sec_interlacement_convergence}, and that are also novel compared to \cite{Aragues-19ACC}.

\section{Notation and Problem Description}
\label{sec_problem_description}

\begin{figure}[tbp]
\begin{center}
\begin{tabular}{c}
\includegraphics[width=0.3\paperwidth]{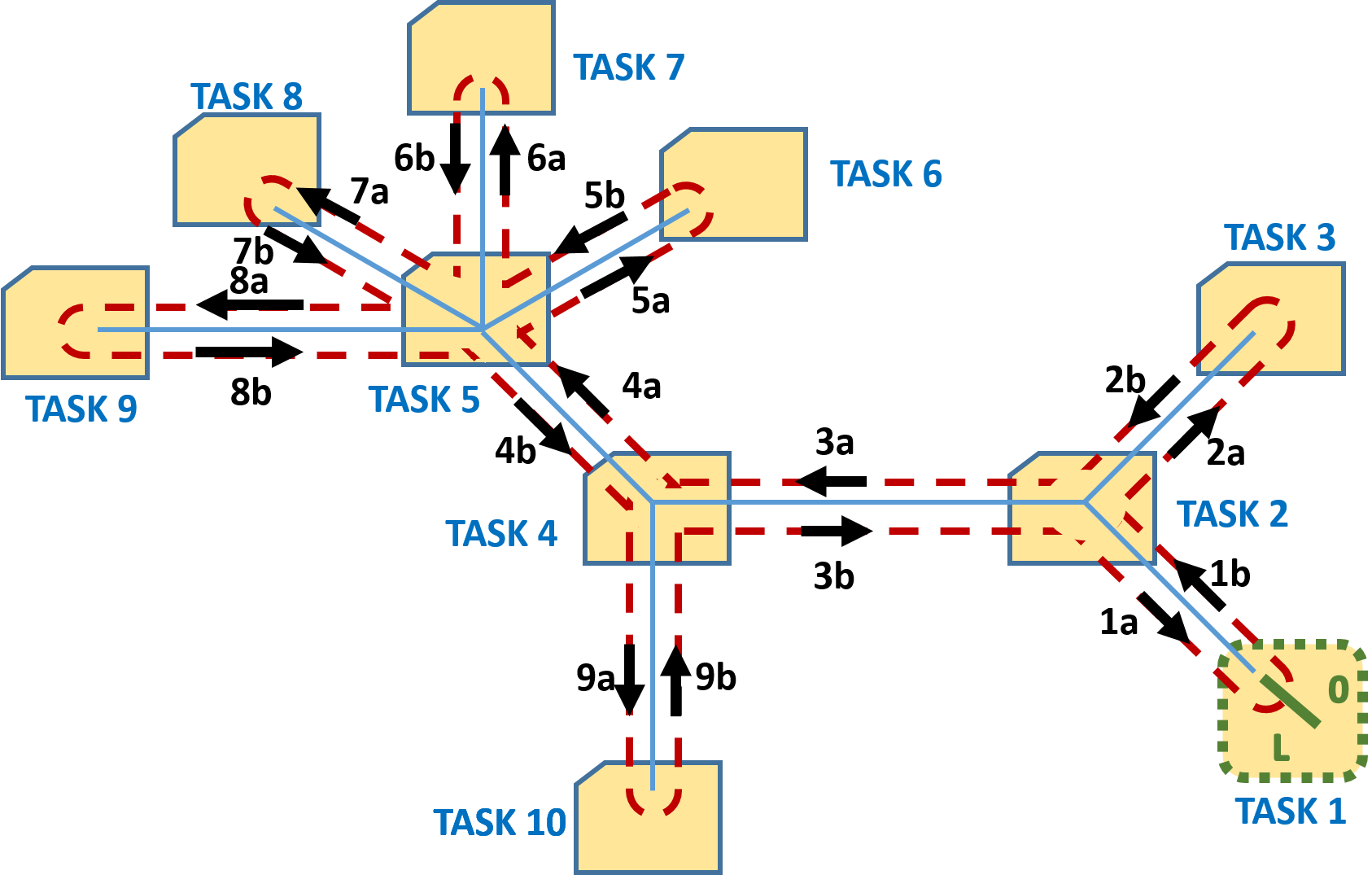}\\
\includegraphics[width=0.3\paperwidth]{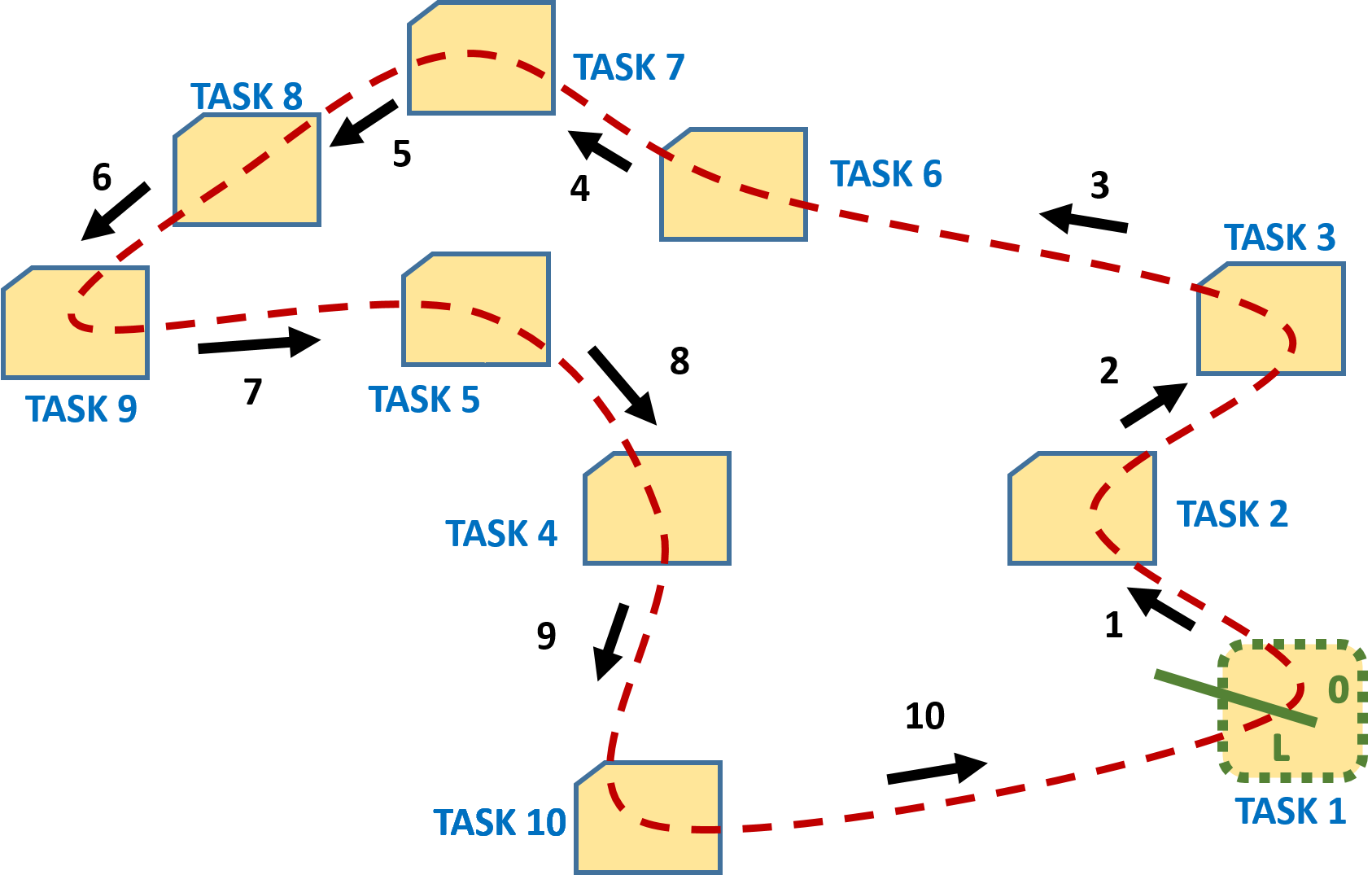}
\end{tabular}
\end{center}
\caption{{\bf(Top)} Example of cycle graph connecting the locations of 10 tasks (orange regions), obtained from a Tree (in blue) with duplicated edges (red dashed). 
We take, e.g., positions 0 and $L$ of the cycle graph at the location of Task 1, between $1a$ and $1b$ (in green). Then, the cycle graph involves edges $1b, 2a, 2b, \dots$ and  finally, $1a$, getting back to the initial position at Task 1. Robots move forward and backward on the cycle graph.
{\bf(Bottom)} Other example of cycle graph (approximate TSP), connecting the task locations. Positions 0 and $L$ of the cycle graph are at the location of Task 1. The edges are $1,2,\dots,10$ (red dashed), getting back to the initial position at Task 1. We do not make any assumptions on the relation between the number of robots $n$  and the number of task locations $l$ and we do not restrict the robots to remain within one specific edge. 
}
\label{fig_tree}
\end{figure}

Assume a team of so called task--robots is in charge of servicing some tasks. Task--robots travel to the different locations of the $l$ tasks placed in an environment as in Fig.~\ref{fig_tree}. To provide task--robots with data exchange capabilities, we place in the area a dedicated team of communicating--robots, that communicate among them and arrive at the task locations periodically, as it is required by classical distributed algorithms such as  distributed averaging, $\max$/$\min$ consensus, or flooding. When a task--robot wants to propagate or get data updates, it just waits at its current task location for a communicating--robot to show up, and then, it exchanges data and moves to its next task location. In this paper, we focus on the behavior of the communicating robots, called from now on \emph{robots}.

A cycle connecting the $l$ task locations is pre--computed or obtained in a centralized fashion, and is available to the communicating--robots. The cycle graph can be built using, e.g., a Minimum--distance Spanning Tree (MST) with duplicated edges \cite{gabriely2001spanning}, or computing an approximate or exact solution of the Traveling Salesman Problem (TSP) \cite{applegate2006traveling}. A location on this cycle  graph is established as position $0$.

There are $n$ communicating--robots (\emph{robots}), with different maximum motion speeds and communication radii, that move forward and backward through the edges of the graph, meeting and exchanging data with their neighbors. We do not make any assumptions on the relation between $n$ and $l$, and also we do not restrict the robots to remain within one specific edge. 

Since every scenario with tasks located on a plane can be transformed into a cycle graph \cite{gabriely2001spanning}, \cite{applegate2006traveling}, from now on, we will no longer consider the underlying structure. We will focus instead on the behavior of the method on the  associated $1D$ cycle  graph (the mapping from this $1D$ cycle graph and the $2D/3D$ scene is commented later in Sec. \ref{sec_experiments}). We let $L$ be the total length of the cycle graph, i.e., the sum of the lengths of the edges that connect the $l$ tasks in the cycle graph. Note that different cycle graphs will give rise to different values of this total length $L$.

We consider $n$ robots moving along the cycle graph. Each  robot $i=1,\dots,n$ 
has a communication radius $r_i\in\mathbb{R}_{\geq 0}$ and a maximum motion speed $v_i\in\mathbb{R}_{> 0}$, and it is assigned a \emph{scalar} $p_i(t)\in\mathbb{R}$, which represents its position in the cycle graph, $p_i(t)\in[0, L]$, for $i=1,\dots,n$. In the  simulations we will represent with a line the robot positions between $0$ and $L$. Due to the 
cyclic structure of the cycle graph, the position $L$ is then equivalent to position $0$.
 Robots cannot move faster than their maximum speed.  At every time instant, each robot $i$ can move forward, backward, or be stopped. This information is represented with the activity 
 $a_i(t)$ and orientation  $o_i(t)$ variables. The activity  
 $a_i(t)\in\{0, 1\}$ represents that robot $i$ is respectively stopped or moving, whereas the     orientation $o_i(t)\in\{-1, +1\}$ represents that robot $i$ moves respectively backward or  forward.  Note that we use two variables $a_i(t)$ and $o_i(t)$ since we want stopped robots to  have an orientation $o_i(t)$ associated to them. This property will be used later in the paper. Robots either move at their maximum speeds or remain stopped, so that 
\begin{align}
\dot{p}_i(t)&= v_i a_i(t) o_i(t).
\label{eq_motion_model}
\end{align} 

\begin{problem}
We assume that the cycle graph cannot be covered by the robots at static positions using their radii $r_i$. Thus, for some periods of time, a task location will remain unconnected (it will have no communicating robot nearby). The aim is to design a strategy for the communicating robots moving and exchanging data on the cycle graph that ensures  the task locations receive the visit of a robot periodically, and that  provides theoretical guarantees on the time elapsed between visits of the robots to the task locations. 
Robots must meet each other so that the information can travel along the  cycle graph (i.e., between all the task locations). The strategy must make use of the capabilities of the robots, which are heterogeneous and have different speeds and communication radii. Moreover, we are interested in providing a solution where robots  exchange data only when they meet in the cycle graph, and that is robust to changes in the capabilities of the robots, i.e., a distributed asynchronous method.
\end{problem}

In the remaining of this section, we explain the notation and definitions used. The proposed  distributed asynchronous method is presented in Sec. \ref{sec_method}, and its properties and performance guarantees are given in Sec. \ref{sec_properties}.

\begin{figure}[tbp]
\begin{center}
\begin{tabular}{cc}
\includegraphics[height=0.1\paperwidth]{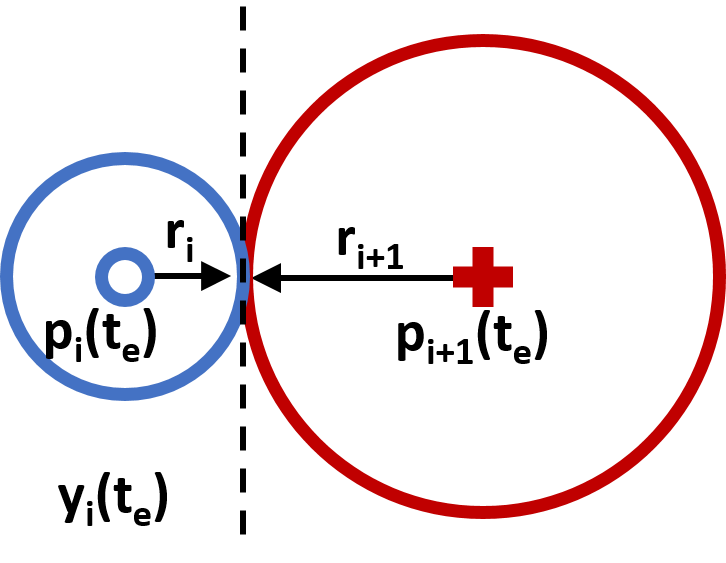}&
\includegraphics[height=0.1\paperwidth]{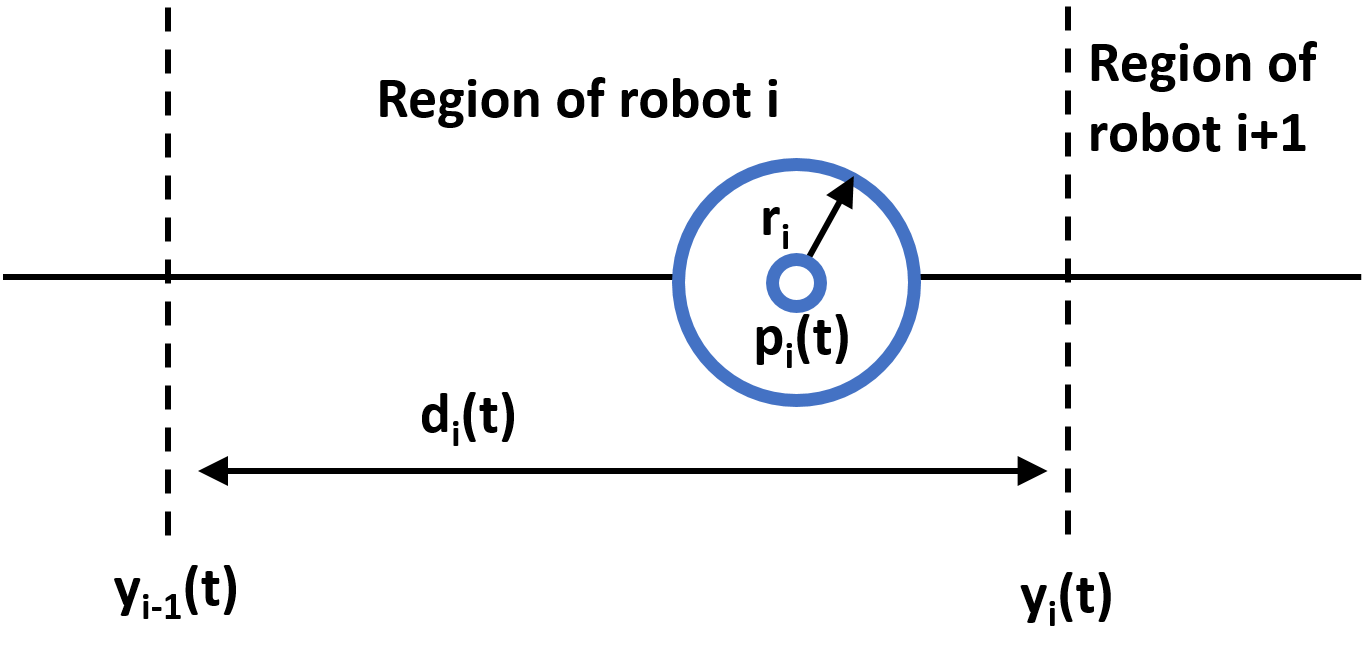}
\end{tabular}
\end{center}
\caption{{\bf Left:} Events like arriving to a boundary and meeting, catching, or discovering a neighbor (described in Sec.~\ref{sec_method}), take place when the communication regions of robots get in touch, or touch the boundary. {\bf Right:} Region associated to robot $i$, length $d_i(t)$ of the region, and position of the boundaries $y_{i-1}(t), y_i(t)$.}
\label{fig_notation}
\end{figure}

Consider the robots on the cycle graph. Each robot $i\in 1,\dots,n$ has two neighbors, its left ($i-1$) and right ($i+1$) neighbor. For the clarity of the presentation, we assume the robot identifiers are sorted according to their position on the cycle graph, from left to right. From now on, $i+1=1$ for $i=n$ , and $i-1=n$ for $i=1$. Between  robots $i$ and $i+1$, for $i= 1,\dots,n$, there is a \emph{boundary} $y_i(t)$ (its computation is explained in Sec.~\ref{sec_method}). 
Each robot $i$ is responsible of the region in the cycle graph within its  boundaries $y_{i-1}(t)$, $y_i(t)$. Robot $i$  moves forward and backward within its region, until its communication zone reaches its boundaries (Fig.~\ref{fig_notation}). When robot $i$ meets its neighbor $i+1$ at the boundary $y_i(t)$ (or neighbor $i-1$ at boundary $y_{i-1}(t)$) at a time $t_{e}$, they can exchange data. We let $d_i(t)$ be the \emph{length} of the region associated to a robot $i$, that depends on its boundaries $y_{i-1}(t)$, $y_{i}(t)$,
\begin{align}
&d_1(t)= y_1(t), ~d_i(t)=y_i(t)-y_{i-1}(t), ~i=2,\dots,n.
\label{eq_d_def}
\end{align}
Note from Fig.~\ref{eq_d_def} that when robot $i$ reaches the boundary $y_i(t)$, its position is $p_i(t)=y_i(t)-r_i$, and when it reaches the boundary $y_{i-1}(t)$, its position is $p_i(t)=y_{i-1}(t) + r_i$. Thus, the \emph{traversing time} $e_i(t)$ that robot $i$ needs to move  between its boundaries (from position $y_i(t)-r_i$ to  $y_{i-1}(t) + r_i$) at maximum speed $v_i$, for $i=1,\dots,n$, is
\begin{align}
e_i(t)&=\frac{y_i(t)-r_i - ( y_{i-1}(t) + r_i)}{v_i} = \frac{d_i(t)-2 r_i}{v_i},
\label{eq_e_def}
\end{align}
where $d_i(t)$ is given by~\eqref{eq_d_def} and, for $i=1$, $y_{i-1}(t)=y_n(t)$.

\subsection{Regions with Common Traversing Times}

In the following discussion, we add a $\star$ symbol to all the variables (region boundaries $y_i^\star$, region lengths $d_i^\star$, and traversing times $e_i^\star$, for $i=1,\dots,n$) to refer to the goal values we want the method to achieve. 
Later, in Sec. \ref{sec_method}, we will present algorithms to compute these variables in a distributed and asynchronous way.
Similar to eqs.~\eqref{eq_d_def} and ~\eqref{eq_e_def}, $y_i^\star$, $d_i^\star$ and $e_i^\star$ are related by:
\begin{align}
d_1^\star&= y_1^\star, ~~~d_i^\star=y_i^\star-y_{i-1}^\star, \mathrm{~for~} i=2,\dots,n,\notag\\
e_i^\star&=(d_i^\star - 2 r_i)/v_i, \mathrm{~for~} i=1,\dots,n.
\label{eq_d_star_def}
\end{align}

\begin{figure}[tbp]
\begin{center}
\includegraphics[height=0.2\paperwidth]{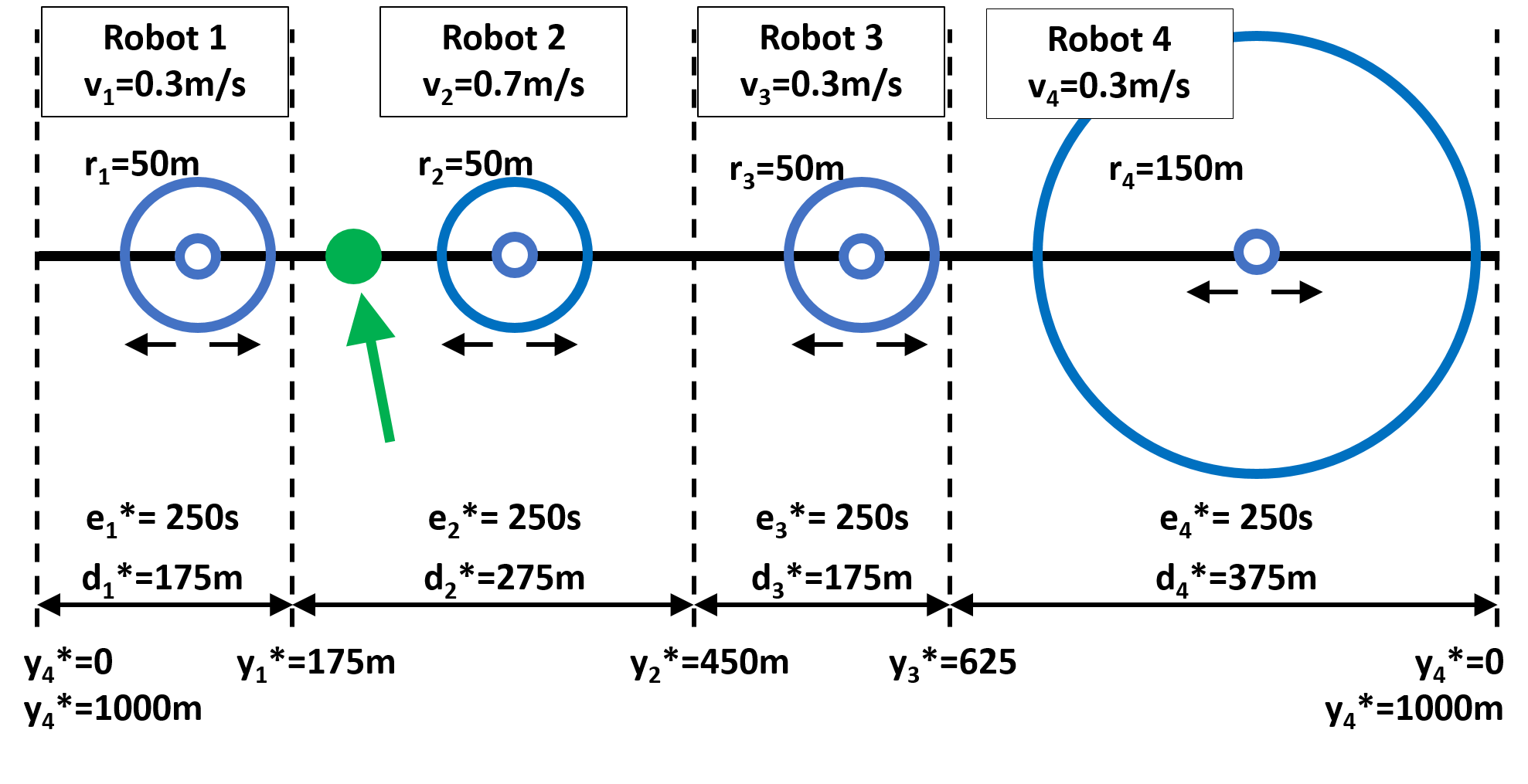}
\end{center}
\caption{Regions with common traversing times. Four heterogeneous robots (blue circles) move forward and backward on a cycle graph with length $L=1000m$. Robots 1 and 3 have maximum speeds and communication radius $v_i=0.3m/s$, $r_i=50m$, $i=1,3$. Robot 2 can move faster ($v_2=0.7m/s, r_2=50m$), and robot 4 has a larger communication radius ($v_4=0.3m/s, r_4=150m$). If we assign regions to them with common traversing times \eqref{eq_t_star_radii}, here $t_\star=e_i=250 s$, $i=1,\dots,n$, and they always move at their maximum speed, then we can ensure that each particular point in the cycle graph (e.g., the green dot), is revisited (Def.~\ref{def_revisiting_time}) every $2 t_\star=500 s$.
}
\label{fig_common_traversing_times_idea}
\end{figure}

Given the cycle graph with length $L$, represented with a line between $0$ and $L$, we want to partition it into $n$ regions and to assign each of these to a robot $i$. The region associated to each robot $i$, for $i=1,\dots,n$, is defined by its boundaries $y_{i-1}^\star, y_i^\star$, and it has associated the length $d_i^\star$. We want each point in the cycle graph to be periodically revisited by the robot $i$ in charge of the associated region. Thus, we are interested in regions that are disjoint, with the only common point being the boundary, and whose union is the cycle graph ($0\dots L$),
\begin{align}
d_1^\star+ d_2^\star + \dots + d_n^\star =L.
\label{eq_d_sum}
\end{align}
An example can be found at Figure~\ref{fig_common_traversing_times_idea}.

In order to take advantage of the maximum speeds $v_i$ and communication radii $r_i, i=1,\dots,n$ of each robot $i$, we want the traversing times $e_i^\star$ employed by each robot $i$ for moving between its boundaries, to be the same, i.e.,  
\begin{align}
t_\star=e_1^\star = \dots e_n^\star,
\label{eq_t_star1}
\end{align}
and we let $t_\star$ be the \emph{common traversing time}. From \eqref{eq_e_def},  \eqref{eq_d_sum}, \eqref{eq_t_star1}, the common  traversing time $t_\star$  is given by:
\begin{align}
&t_\star=\frac{d_1^\star - 2 r_1}{v_1} = \frac{d_2^\star - 2 r_2}{v_2} =\dots = \frac{d_n^\star - 2 r_n}{v_n}, \notag\\
&(v_1+\dots+v_n)t_\star=d_1^\star+\dots+d_n^\star - 2r_1-\dots-2r_n,\notag\\
&t_\star = (L  - 2 \mathop{\sum}_{i=1}^n r_i)/(v_1+v_2+\dots+v_n).
\label{eq_t_star_radii}
\end{align}
Having smaller environments, more robots, faster, or with larger communication radii,  produce lower common traversing times (Fig. \ref{fig_common_traversing_times_idea}).

Before continuing, we point out some facts about $t_{\star}$ \eqref{eq_t_star_radii}. Suppose that, instead of $n$ robots, there was a single entity traversing the cycle graph,  comprising the capabilities of all the robots: with maximum speed equal to $v_1+\dots+v_n$ and communication radius equal to $r_1+\dots+r_n$ (and diameter twice this  quantity). In order to traverse a cycle graph with total length $L$, this single entity would employ a time equal to \eqref{eq_t_star_radii}. 
In order to achieve this performance with a multi--robot team, robots must be correctly organized, as we propose in this paper. Note that the traversing times of all the robots must equal $t_{\star}$. Otherwise, if a robot has a shorter traversing time, other robots will necessarily have longer traversing times, since they have to cover longer regions. Thus, we would not take full advantage of their capabilities. 

From \eqref{eq_d_star_def}, \eqref{eq_t_star_radii}, the length $d_i^\star$ of the  region associated to robot $i$, for $i=1,\dots,n$, is:
\begin{align}
d_i^\star &= v_i t_\star + 2 r_i = \frac{v_i (L  - 2 \mathop{\sum}_{j=1}^n r_j)}{v_1+v_2+\dots+v_n} + 2 r_i,
\label{eq_d_star_radii}
\end{align}
and the boundary position $y_i^\star =d_1^\star+\dots+d_i^\star$, for $i=1,\dots,n$, obtained by summing up the region lengths $d_{j}^\star$, is:
\begin{align}
y_i^\star&= \mathop{\sum}_{j=1}^i \frac{v_j (L- 2 \mathop{\sum}_{j'=1}^n r_{j'})}{v_1+v_2+\dots+v_n} + 2r_j,
\label{eq_optimal_boundary}
\end{align}
where $y_n^\star$ gives $L$ as expected. 

\begin{definition}[Revisiting times]: 
We define the \emph{revisiting time} as the time required for a robot to visit a particular point in the cycle graph, arriving back at the point with the same orientation. 
\label{def_revisiting_time}
\hfill \IEEEQEDhere
\end{definition}

Note that the \emph{revisiting time} $t_{\mathrm{rev}}$ includes:
\begin{itemize}
\item $2 e_i(t)$ to traverse the robot region in both directions at maximum speed, getting back to the original point with the same orientation, plus
\item additional times where the robot is e.g., stopped.
\end{itemize}
In the definition of the \emph{revisiting time} $t_{\mathrm{rev}}$ we impose that the orientation of the robot must be the same, in order to represent that robot $i$ has completed traversing its associated region (move in one direction until it reaches its boundary, reverse and move to the opposite boundary, reverse and move back until the starting position). Thus, the condition on the orientation and position  of the robot being the same represents that robot $i$ carries out fresh data from both neighbors $i-1$ and $i+1$, that now is available to the revisited point.

Note that the optimal boundaries for each robot ~\eqref{eq_optimal_boundary} could be pre--computed in an off--line fashion (centralized alternative). In this paper, we consider that the task locations are static or slowly changing, whereas the team of  communicating--robots is more dynamic. Thus, off--line centralized computation is reasonable for the  cycle graph, since it only depends on the task locations. On the other hand, we are interested  in solutions that do not require knowing and keeping track of the characteristics of all the  involved communicating--robots (the total amount  $n$ of robots, their communication radii $r_i$, their speeds $v_i$, and their ordering in the cycle graph, with the associated cost linear in $n$), and that can self adapt to changes in the team (e.g., robots increasing / decreasing their  communication radius \cite{Hong-19ACC},  \cite{Sabattini_trCyber2015} or their maximum speeds). Thus, we propose a distributed solution, where each robot computes asymptotically its boundaries \eqref{eq_optimal_boundary} using only local information. This distributed method is presented in Sec. \ref{sec_method}.

Note also that, if robots were assigned regions with common traversing times $t_\star$ ~\eqref{eq_t_star_radii} and they never stopped, then each point would have a revisiting time equal to $t_{\mathrm{rev}}=2t_\star$. However, robots can only exchange data when they meet, and this requires some additional coordination that may make the performance degrade. In this paper, we propose a method where, under some conditions, robots achieve $t_{\mathrm{rev}}=2t_\star$. We provide a thorough analysis of the performance degradation when these conditions are not satisfied.

\section{Intermittent Connectivity Maintenance with Heterogeneous Robots}
\label{sec_method}

Robots run the distributed asynchronous algorithm presented in this section for meeting intermittently, and for computing their boundaries $y_i(t)$. Later, in Sections ~\ref{sec_properties} to \ref{sec_interlacement_convergence}, we discuss the  properties of the method, and the relation between the  boundaries $y_i(t)$ computed in a distributed way and the boundaries $y_i^{\star}$ that could be obtained if all the information from all the robots was known by a central unit.

The method roughly consists of each robot $i$ moving until it reaches or defines a boundary, waiting at this boundary until it meets with its neighbor, updating their data, and then moving to its other boundary and repeating the process. We distinguish between the following behaviors for the robots:
\begin{itemize}
\item Participating in an \emph{event}: events are associated with an event time 
 $t_{e_1}, t_{e_2}, t_{e_3},\dots $, or generically $t_{e}$, they affect at most two neighbors, and they modify their values of the activity $a_i(t_e)$, orientation $o_i(t_e)$, and boundary $y_i(t_e)$. 
\item \emph{Between events}: robot positions $p_i(t)$ evolve according to \eqref{eq_motion_model}. Robots may be active ($a_i(t)=1$), moving from their current position until they arrive to or define a boundary ($a_i(t)=1$), or inactive ($a_i(t)=0$), waiting at a boundary.
\end{itemize}
Along the section, we define the types of events, and how robots react to them. We make the following assumptions:

\begin{assumption}[A1, A2, A3]
\label{ass_method}
$(A1)$ Robots $n$ and $1$ have a fixed boundary $y_n(t)=L$ for all $t\geq 0$, placed at position $L$ (equivalently, due to the cycle structure, at position $0$).  
$(A2)$ $o_i(0)\neq o_j(0)$ for at least a pair of robots $i,j, i\neq j$.
$(A3)$ Robots $i=1,\dots,n$ start placed at positions $p_i(0)\in[0, L]$ so that their  communication  zones do not overlap, with initial orientations $o_i(0)$ satisfying $(A2)$, and all are active  $a_i(0)=1,$ for all  $i=1,\dots,n$. 
\end{assumption}

The proposed algorithm works as follows. 

\begin{algo}[Discovery and Catch]
\label{algo_discovery}
Initially, robots do not know  their boundaries $y_i(0), i=1,\dots,n-1$ but $y_n(0)=L$ from Assumption $(A1)$. Robots move to discover their neighbors and set an initial value for their boundaries. There are two events associated:

\noindent \emph{Discovery} event: Robots $i$ and $i+1$ are moving,  ($a_i(t)=1$  and $a_{i+1}(t)=1$), with robot $i$ moving forward and robot $i+1$ moving  backward, $o_i(t)>0, o_{i+1}(t)<0$, and they do not know $y_i(t)$. The discovery happens when their communication regions get in touch at $t_{e}\geq t$ (Fig.~\ref{fig_notation}),
\begin{align}
p_i(t_{e})+r_i = p_{i+1}(t_{e}) - r_{i+1}.
\label{eq_meeting_discovery}
\end{align}
At the \emph{discovery}, the involved robots $i$ and $i+1$ initialize their common boundary $y_i(t)$ with this discovery position, they reverse their orientations and move in opposite directions, as follows:
\begin{align}
y_i(t_{e}^+)=p_i(t_{e})+r_i, ~~o_i(t_e^+)=-1, ~o_{i+1}(t_e^+)= +1.
\label{eq_after_discovery}
\end{align}

\noindent \emph{Catch} event: Robots $i$ and $i+1$ do not know $y_i(t)$. 
The catcher is robot $i$ when it is active $a_i(t)=1$, both $i$ and $i+1$ are oriented forward $o_i(t)>0, o_{i+1}(t)>0$, and robot $i+1$ is stopped $a_{i+1}(t)=0$  or moving slower $a_{i+1}(t)=1, v_{i+1} < v_i$. (Equivalently, the catcher is robot $i+1$ when $a_{i+1}(t)=1, o_i(t)<0, o_{i+1}(t)<0$, and robot $i$ is stopped $a_{i}(t)=0$ or moving slower $a_{i}(t)=1, v_{i} < v_{i+1}$). 

The catch happens when their communication regions get in touch at $t_e\geq t$ as  in eq.~\eqref{eq_meeting_discovery} and Fig.~\ref{fig_notation}. At the \emph{catch}, robots $i$ and $i+1$ initialize their  common boundary $y_i(t)$, the catcher robot (e.g., robot $i$) remains waiting at this boundary, and the caught robot $i+1$ moves or keeps on moving towards the opposite boundary,
\begin{align}
y_i(t_{e}^+)&=p_i(t_{e})+r_i, &a_i(t_e^+)&= 0, &a_{i+1}(t_e^+)&= 1.
\label{eq_after_catch}
\end{align}
\end{algo}

Note that the updates due to the events \eqref{eq_after_discovery}, \eqref{eq_after_catch} are  designed so that the number of positive and negative robot orientations is kept. For the \emph{catch} event, this requires the catcher to remain stopped at the boundary (instead of reversing its orientation, as it happens for the \emph{discovery}).

During the first time instants, some robots may be discovering and catching  neighbors (Algorithm~\ref{algo_discovery}), and others may have already discovered them. Once robot $i$ has discovered both its neighbors, it knows $y_{i-1}(t), y_i(t)$, and it moves within these boundaries, updating them, and acting from then on according to the following algorithm. 

\begin{algo}[Arrivals and Meetings]
\label{algo_main}
Robot $i$ executes this algorithm only when it already knows its boundaries  $y_{i-1}(t), y_i(t)$ (otherwise, it moves to discover or catch its neighbor as per Algorithm~\ref{algo_discovery}). 
In this phase, the events that can take place are the following:

\noindent \emph{Arrival} event: Robot $i$ moving backward $a_i(t)=1$, $o_i(t)<0$ (or moving  forward $a_i(t)=1$, $o_i(t)>0$) arrives at its $y_{i-1}(t)$ boundary (or at $y_{i}(t)$ respectively) at a time $t_{e}\geq t$ when its communication region touches the boundary (Fig.~\ref{fig_notation}), i.e.,
\begin{align}
&y_{i-1}(t_{e})=p_i(t_{e})-r_i, ~(\mathrm{resp.~} y_{i}(t_{e})=p_i(t_{e})+r_i)
\label{eq_arrival}
\end{align}
After an \emph{arrival} to a boundary, robot $i$ waits at the boundary,
\begin{align}
a_i(t_e^+) &= 0.
\label{eq_after_arrival}
\end{align}
The event is \emph{arrival} if robot $i$ arrives to the boundary and there is no neighbor waiting there. Otherwise, it is a \emph{meeting}  event.

\noindent \emph{Meeting} event: Robots $i$ and $i+1$ meet at time $t_{e}\geq t$  when both of them arrive at their common boundary $y_i(t)$,
\begin{align}
y_{i}(t_{e}) &= p_i(t_{e})+r_i = p_{i+1}(t_{e}) - r_{i+1}.
\label{eq_meeting_boundary}
\end{align}
At the \emph{meeting}, robots $i$, $i+1$ update their common boundary:

{\small
\begin{align}
y_i(t_e^+)=\frac{v_{i+1}(y_{i-1}(t_e)+2r_i) + v_i(y_{i+1}(t_e)- 2r_{i+1})}{v_i + v_{i+1}}.
\label{eq_interMeetingPoint_update_radii}
\end{align}}

\noindent Then, robots reverse their orientations and get active,

{\small
\begin{align}
o_i(t_e^+)=-1, o_{i+1}(t_e^+)=+1, a_i(t_e^+)=1, a_{i+1}(t_e^+)= 1.
\label{eq_after_meeting}
\end{align}}

\noindent In fact, only $i$ or $i+1$ (the first one that arrived to the boundary) should be inactive. If both arrivals take place at the same time, we establish an order, e.g., to start with the left robot. 

After the \emph{meeting} and the updates \eqref{eq_interMeetingPoint_update_radii}, \eqref{eq_after_meeting}, robots $i$, $i+1$ move away from each other, towards  their opposite boundary. E.g., robot $i$ moves towards the common boundary $y_{i-1}(t)$ with its neighbor $i-1$. Note that robots $i$ and $i-1$ must have the same value for $y_{i-1}(t)$, since it can only be updated when both $i$ and $i-1$ meet, and thus it cannot have been changed in the meantime. When robot $i$ gets to the $y_{i-1}(t)$ boundary, a new \emph{arrival} or \emph{meeting}  takes place.
\end{algo}

\begin{remark}[Execution using information of times]
In the above algorithm, the update of the robot regions is performed by taking into account the positions of the boundaries $y_i(t_e)$ \eqref{eq_interMeetingPoint_update_radii}. As we will see 
 later, in the Proof of Prop.~\ref{prop_weightedConsensus} (Appendix \ref{appendices_prop_weightedConsensus}), the previous algorithm can be  equivalently run by  considering traversing times instead, i.e., $e_i(t_e)$:
\begin{align}
&e_{i}(t_e^+)=e_{i}(t_e) + \frac{\epsilon_i}{v_{i}} (e_{i+1}(t_e) -e_{i}(t_e) ), \mathrm{~} \epsilon_i = \frac{v_i v_{i+1}}{v_i+v_{i+1}}, \notag\\
&e_{i+1}(t_e^+)=e_{i+1}(t_e) - \frac{\epsilon_i}{v_{i+1}} (e_{i+1}(t_e) - e_{i}(t_e)). \label{eq_region_time_e_update_2}
\end{align}
 Robots, instead of moving until they touch the boundary, would move at their maximum speed $v_i$ for $e_i(t_e^+)$ time.
\end{remark}

\begin{remark}[On Assumption \ref{ass_method}]

In the next sections, we will analyze the performance of the method under Assumption \ref{ass_method}. Note that, in fact, Assumption \ref{ass_method}(A1) can always be ensured: robots $1$ and $n$ do not need to know they are the first and the last ones. During the discovery and catch (Algorithm \ref{algo_discovery}), if a robot moving  backwards gets to the position $p_i(t) = 0+r_i$   (i.e., it arrives to a boundary placed  at the position 0 in the cycle graph), then it records this boundary and remains waiting at this boundary (equivalently, a robot moving forward and arriving to the position $L$ in the cycle graph). Assumption  \ref{ass_method}(A3) is required for simplifying the analysis, although in practical setups it can be relaxed so that robots navigate to reach the cycle graph.
\end{remark}

\begin{algorithm}[!ht]                      
\caption{\emph{Discovery and Catch (Alg.~\ref{algo_discovery})} - Robot $i$ ($v_i,r_i,p_i(0), o_i(0)$)}          
\label{alg_roboti_discovery}                           
\begin{algorithmic}[1]                
\State Initialize empty boundaries ($y_{i-1}(0):=<>$, $y_{i}(0):=<>$)
\State Get active ($a_i(0):=1$)
\While{~either $y_{i-1}(t)$ or $y_{i}(t)$ are empty}
\State Move according to \eqref{eq_motion_model} until an event occurs:

\Switch{event}
\State\LeftComment{0}{Events for forward behavior ($o_i(t)=1$)}
\Case \emph{discovery} of neighbor $i+1$:
\State Initialize boundary ($y_{i}(t):=p_i(t)+r_i$)
\State Reverse orientation ($o_i(t):=-1$)
\EndCase
\Case \emph{catch} of neighbor $i+1$ (or \emph{arrival} to $y_n=L$):
\State Initialize boundary ($y_{i}(t):=p_i(t)+r_i$)
\State Wait at boundary $y_i(t)$ ($a_i(t):=0$)
\State Resume when neighbor $i+1$ arrives to $y_i(t)$
\State Get active ($a_i(t):=1$)
\State Reverse orientation ($o_i(t):=-1$)
\EndCase
\Case \emph{catch}: robot $i$ is caught by neighbor $i-1$:
\State Initialize boundary ($y_{i-1}(t):=p_i(t)-r_i$)
\State Keep orientation ($o_i(t)=1$)
\EndCase
\State\LeftComment{0}{Events for backward behavior ($o_i(t)=-1$)}
\Case \emph{discovery} of neighbor $i-1$:
\State Initialize boundary ($y_{i-1}(t):=p_i(t)-r_i$)
\State Reverse orientation ($o_i(t):=+1$)
\EndCase
\Case \emph{catch} of neighbor $i-1$ (or \emph{arrival} to $y_n=L$):
\State Initialize boundary ($y_{i-1}(t):=p_i(t)-r_i$)
\State Wait at boundary $y_{i-1}(t)$ ($a_i(t):=0$)
\State Resume when neighbor $i-1$ arrives to $y_{i-1}(t)$
\State Get active ($a_i(t):=1$)
\State Reverse orientation ($o_i(t):=+1$)
\EndCase
\Case \emph{catch}: robot $i$ is caught by neighbor $i+1$:
\State Initialize boundary ($y_{i+1}(t):=p_i(t)+r_i$)
\State Keep orientation ($o_i(t)=-1$)
\EndCase
\EndSwitch
\EndWhile
\State Run \textit{Arrivals and Meetings} algorithm (Alg.~\ref{alg_roboti_main})
\end{algorithmic}
\end{algorithm}

\begin{algorithm}[!ht]                      
\caption{\emph{Arrivals and Meetings (Alg.~\ref{algo_main})} - Robot $i$ ($v_i,r_i,p_i(t), o_i(t), a_i(t), y_{i-1}(t), y_{i}(t)$)}          
\label{alg_roboti_main}                           
\begin{algorithmic}[1]                
\While{true (run for ever)}
\State Move according to \eqref{eq_motion_model} until an event occurs:

\Switch{event}
\State\LeftComment{0}{Events for forward behavior ($o_i(t)=1$)}
\Case \emph{arrival} to boundary $y_i(t)$
\If{neighbor $i+1$ is not yet at $y_i(t)$}
\State Wait at boundary $y_i(t)$ ($a_i(t):=0$)
\State Resume when neighbor $i+1$ arrives to $y_i(t)$
\EndIf
\State\LeftComment{0}{\emph{Meeting} event}
\If{$i<n$ ($y_n=L$ fixed, Asm.~\ref{ass_method})}
\State Send $v_i,r_i,y_{i-1}(t)$ to neighbor $i+1$
\State Receive $v_{i+1},r_{i+1},y_{i+1}(t)$ from $i+1$
\State Update $y_i(t)$ with eq. \eqref{eq_meeting_boundary}
\EndIf
\State Get active ($a_i(t):=1$)
\State Reverse orientation ($o_i(t):=-1$)
\EndCase
\State\LeftComment{0}{Events for backward behavior ($o_i(t)=-1$)}
\Case \emph{arrival} to boundary $y_{i-1}(t)$
\If{neighbor $i-1$ is not yet at $y_{i-1}(t)$}
\State Wait at boundary $y_{i-1}(t)$ ($a_i(t):=0$)
\State Resume when neig. $i-1$ arrives to $y_{i-1}(t)$
\EndIf
\State\LeftComment{0}{\emph{Meeting} event}
\If{$i>1$ ($y_n=L$ fixed, Asm.~\ref{ass_method})}
\State Send $v_i,r_i,y_{i}(t)$ to neighbor $i-1$
\State Receive $v_{i-1},r_{i-1},y_{i-2}(t)$ from $i-1$
\State Update $y_{i-1}(t)$ with eq. \eqref{eq_meeting_boundary}
\EndIf
\State Get active ($a_i(t):=1$)
\State Reverse orientation ($o_i(t):=+1$)
\EndCase
\EndSwitch
\EndWhile
\end{algorithmic}
\end{algorithm}

For clarity, we include in Algorithms~\ref{alg_roboti_discovery} and ~\ref{alg_roboti_main} the pseudo--code instructions 
that are run by each robot $i$ participating in Alg. \ref{algo_discovery} (Discovery and  Catch) and Alg. \ref{algo_main} (Arrivals and Meetings). 
First, robot $i$ runs Alg.~\ref{alg_roboti_discovery} (Discovery and Catch), 
using its maximum speed $v_i$ and radius $r_i$, its initial position on the cycle graph 
$p_i(0)$ and its initial orientation $o_i(0)$. Once it has discovered its two boundaries, it proceeds with the main algorithm (Arrivals and Meetings, Alg. \ref{alg_roboti_main}), 
using its maximum speed $v_i$, radius $r_i$, 
current position on the cycle graph $p_i(t)$, orientation $o_i(t)$, activity $a_i(t)$  and the current values for the boundaries ($y_{i-1}(t), y_i(t)$).

Observe the low complexity of the proposed method (Algs. ~\ref{alg_roboti_discovery} and  \ref{alg_roboti_main}), which is constant in memory, computation,and data exchange at  each robot encounter.

\section{Main Results}
\label{sec_properties}

Here we state the main properties of the method. The proofs of Theorems~\ref{th_common_traversing_times}, ~\ref{th_performance_balanced} and  \ref{th_performance_unbalanced} are given in Appendix \ref{appendices_ths}, and they depend on several properties presented in Sections~\ref{sec_convergence_common_traversing_times},~\ref{sec_interlaced} and \ref{sec_interlacement_convergence}.

\begin{theorem}[Convergence to common traversing times]
\label{th_common_traversing_times}
Consider that robots run Algorithms \ref{algo_discovery}, \ref{algo_main}  
in Section~\ref{sec_method} under Assumptions~\ref{ass_method}. 
Then, the traversing times $e_i(t)$, region lengths $d_i(t)$, and boundaries $y_i(t)$ (eqs.~\eqref{eq_e_def}, \eqref{eq_d_def}, \eqref{eq_interMeetingPoint_update_radii}) asymptotically converge to the goal values $t_\star$, $d_i^\star$, $y_i^\star$ in eqs.~\eqref{eq_t_star_radii}, \eqref{eq_d_star_radii}, \eqref{eq_optimal_boundary}, for $i=1,\dots,n$.
\end{theorem}
\begin{IEEEproof}
See Appendix \ref{appendices_ths}.
\end{IEEEproof}

The performance achieved depends on the amount of robots moving with positive and negative orientations.

\begin{definition}[Balanced and Unbalanced orientations]
\label{def_balanced_unbalanced}
Let $n_+$ and $n_-$ be the number of robots initially moving with positive ($o_i(0)>0$) and negative  orientations  ($o_i(0)<0$), with $n=n_+ + n_-$, and let $n_{bal}$ be $n_{bal}= \min\{n+,n_-\}$. Robot orientations are \emph{balanced} when $n_{bal}=n_+=n_-=n/2$ (note that $n$ must be even in this case), and they are \emph{unbalanced} otherwise. Without loss of generality, in the paper we consider that there are more robots with  positive orientations so that $n_+ \geq n_-$ (all the discussions apply equivalently to the opposite case).
\hfill \IEEEQEDhere
\end{definition}

\begin{theorem}[Performance for Balanced Orientations]
\label{th_performance_balanced}
A robot team with balanced orientations (Def.~\ref{def_balanced_unbalanced}) running 
Algorithms \ref{algo_discovery}, \ref{algo_main}  
in Section~\ref{sec_method} 
under Assumptions~\ref{ass_method} and using the common traversing times $e_i(t)=t_\star$ for $i=1,\dots,n$ \eqref{eq_t_star_radii}, converges to a  configuration where the $n$ robots perform all their $n/2$  meetings simultaneously, and with \emph{revisiting time} $t_{\mathrm{rev}}$ (Def.~\ref{def_revisiting_time}) given by
\begin{align}
t_{\mathrm{rev}}&=2 t_\star.
\label{eq_th_perfomance_balanced}
\end{align}
\end{theorem}
\begin{IEEEproof}
See Appendix \ref{appendices_ths}.
\end{IEEEproof}

\begin{theorem}[Performance for Unbalanced Orientations]
\label{th_performance_unbalanced}
A robot team with unbalanced orientations (Def.~\ref{def_balanced_unbalanced}) running 
 Algorithms \ref{algo_discovery}, \ref{algo_main}  
in Section~\ref{sec_method} under Assumptions~\ref{ass_method} and using the common traversing times $e_i(t)=t_\star$ for $i=1,\dots,n$ \eqref{eq_t_star_radii}, converges to a configuration where, at every round, there are $n_{bal}$ meetings involving $2 n_{bal}$ robots. Each pair of robots meet at their common boundary $n_{bal}$ times every $n$ rounds. The \emph{revisiting  time} $t_{\mathrm{rev}}$ (Def.~\ref{def_revisiting_time}), averaged along $n_{bal}$ meetings, is given by
\begin{align}
t_{\mathrm{rev}}&= t_\star n / n_{bal}.
\label{eq_th_perfomance_unbalanced}
\end{align}
\end{theorem}
\begin{IEEEproof}
See Appendix \ref{appendices_ths}.
\end{IEEEproof}

In the proposed algorithm, the task locations are not connected at all times. Instead, they are disconnected most of the time, and they are visited from time to time by communicating robots. The interest of Theorems \ref{th_performance_balanced} and  \ref{th_performance_unbalanced} is that they provide theoretical guarantees on the  elapsed time that a particular location on the cycle graph (for instance, a task location) will remain disconnected in the configuration asymptotically achieved by the robot team. The task location will receive in average two visits of a communicating robot every $t_{\mathrm{rev}}$ time (the value is exact for balanced orientations, Th. \ref{th_performance_balanced}). 
Note also that these performance metrics only depend on the number of robots with  balanced orientations (Def.~\ref{def_balanced_unbalanced}) and on the common traversing time $t_\star$ \eqref{eq_t_star_radii}, which depends on the total length of the cycle graph $L$ and on the maximum speeds and communication radii of all the involved robots.

The performance metrics given in Theorems \ref{th_performance_balanced} and \ref{th_performance_unbalanced} will be clearer in Sec. \ref{sec_interlaced}.

\begin{remark}
Observe from Theorems \ref{th_common_traversing_times}, \ref{th_performance_balanced} and \ref{th_performance_unbalanced} that one of the main strengths of the solution we propose, is that it does not depend on the robot IDs or their initial positions, i.e., the same common traversing times and revisiting times are obtained regardless of the robot initial ordering or initial positions in the cycle graph.
\end{remark}

In Section \ref{sec_experiments} we present simulations carried out on the $1D$ cycle graph, we explain the mapping between the positions on this $1D$  cycle graph and the positions on $2D$ or $3D$ environments, and we present additional simulations on  $2D$ environments using differential--drive ground robots. 
In Sec. \ref{sec_convergence_common_traversing_times}, \ref{sec_interlaced}, \ref{sec_interlacement_convergence} and in the appendices, we present several  theoretical results which are required in order to prove Theorems~\ref{th_common_traversing_times}, \ref{th_performance_balanced}, \ref{th_performance_unbalanced}. For clarity, the analysis in these sections is performed considering the $1D$ cycle graph.

\section{Convergence to Common Traversing Times}
\label{sec_convergence_common_traversing_times}

Here, we discuss the proof of Theorem~\ref{th_common_traversing_times}. It relies on auxiliary results from \cite{Aragues-19ACC}, which are included here to make the manuscript self--contained. Some of these results are also used later in Sections \ref{sec_interlaced} and \ref{sec_interlacement_convergence}. 
To prove Theorem~\ref{th_common_traversing_times}, first, we rewrite eq.  \eqref{eq_interMeetingPoint_update_radii} in terms of the traversing times $e_i(t)$  \eqref{eq_e_def} and show that it is an asynchronous weighted consensus method~\cite{olfati2007consensus}, \cite{zhang2011distributed}. We prove its convergence in Proposition~\ref{prop_weightedConsensus}, assuming that the set of communication graphs that \emph{occur infinitely often} \cite{elsner1990convergence} \cite{xiao2005scheme} are jointly connected. An event occurs infinitely often \cite{elsner1990convergence,xiao2005scheme,kantaros2019temporal,KhodayiKantarosTRO2019, GuoTRO2018} if, considering an infinite sequence of events, the particular event in the sequence holds true for an infinite number of indices. 
Then, in Proposition~\ref{prop_jointconn}, we prove that the set of communication graphs that occur infinitely often are indeed jointly connected.

\begin{proposition}\emph{(Weighted consensus on traversing times \cite[Prop.~5.1]{Aragues-19ACC}):}
\label{prop_weightedConsensus}
Assume that algorithm~\eqref{algo_main} gives rise to a network in which the set of communication graphs that occur infinitely often are jointly connected.

 Then, the traversing times $e_i(t)$, region lengths $d_i(t)$, and boundaries $y_i(t)$ (eqs.~\eqref{eq_e_def}, \eqref{eq_d_def}, \eqref{eq_interMeetingPoint_update_radii}) asymptotically converge to the goal values $t_\star$, $d_i^\star$, $y_i^\star$ in eqs.~\eqref{eq_t_star_radii}, \eqref{eq_d_star_radii}, \eqref{eq_optimal_boundary}, for $i=1,\dots,n$.
\end{proposition}
\begin{IEEEproof}
See Appendix \ref{appendices_prop_weightedConsensus}.
\end{IEEEproof}

We give some intermediary results to prove that, under our algorithm, the 
 set of communication graphs that occur infinitely often are jointly connected. 
 In our discussion, we focus on Algorithm~\ref{algo_main} and consider that each   robot $i$ has run the Discovery and Catch phase (Algorithm~\ref{algo_discovery}) and thus has set  an initial value for both boundaries $y_i(t), y_{i+1}(t)$. Note that Algorithm~\ref{algo_discovery} is only run during the first time instants and, after that, robots always run Algorithm~\ref{algo_main}.

\begin{lemma}[(Active){\cite[Lemma 5.1]{Aragues-19ACC}}]
\label{lemma_STmove}
Robots are active ($a_i(t)=1$, Section~\ref{sec_method}) during a bounded time and, after  that, an \emph{arrival} or a \emph{meeting} event always occurs.
\end{lemma}
\begin{IEEEproof}
See Appendix \ref{appendices_lemmas_weigthedConsensus}.
\end{IEEEproof}

This observation allows us to focus on the behavior of the discrete asynchronous version of the method.

\begin{definition}[Discrete asynchronous behavior]
\label{def_discreteAsynchVersion}
The discrete asynchronous version of the method (Algorithm~\ref{algo_main}), includes only the event times  $t_{e_1}, t_{e_2}, t_{e_3},\dots $. Each robot $i$ is always placed at one of its boundaries \eqref{eq_arrival}, $p_i(t_{e})\in\{y_{i-1}(t_{e})+r_i , y_{i}(t_{e})-r_i \}$. 
The states $y_i(t_{e})$, $o_i(t_{e})$, $a_i(t_{e})$, change due to   \emph{meeting} events \eqref{eq_interMeetingPoint_update_radii}, \eqref{eq_after_meeting} (equivalently, $d_i(t_{e})$, $e_i(t_{e})$  \eqref{eq_d_def}, \eqref{eq_e_def}). After a meeting between robots $i$, $i+1$ at time $t_{e}$, two arrival events take place in the future: 
\begin{align}
t_{e'}&=t_{e}+e_i(t^+_{e}), &&p_i(t_{e'}^+)=y_{i-1}(t_{e})+r_i, ~\mathrm{and}\notag\\
t_{e''}&=t_{e}+e_{i+1}(t^+_{e}), &&p_{i+1}(t_{e''}^+)=y_{i+1}(t_{e})-r_{i+1}.
\label{eq_discreteAsynchVersion_event_time}
\end{align}
\hfill\IEEEQEDhere
\end{definition}

\begin{lemma}[Discrete Asynchronous Behavior]
\label{lemma_discrete_asynchronous_behavior}
Assume all robots have finished the Discovery and Catch phase (Algorithm~\ref{algo_discovery}). Then 
the discrete asynchronous behavior (Def.~\ref{def_discreteAsynchVersion}) and Algorithm~\ref{algo_main} have the same:
($i$) Events (arrivals and meetings) and event times $t_{e_1}, t_{e_2}, t_{e_3},\dots$; 
($ii$) States $p_i(t_e)$, $o_i(t_e)$, $a_i(t_e)$ at the event time $t_e$, for the robot or robots involved in the event.
\end{lemma}
\begin{IEEEproof}
See Appendix \ref{appendices_lemmas_weigthedConsensus}.
\end{IEEEproof}

Note that at a time event $t_e$, the states of the robots \emph{not involved} in the meeting will differ between Def.~\ref{def_discreteAsynchVersion} and Algorithm~\ref{algo_main}, since in the first one it is as if they were still on the boundary, whereas in the second one they are currently moving. However,  the event only depends on the robots involved and not on the remaining ones. Thus, Lemma~\ref{lemma_discrete_asynchronous_behavior} holds, and we can use the representation in Def.~\ref{def_discreteAsynchVersion} to study the method in a simpler way.

\begin{lemma}[Properties{\cite[Lemma 5.2]{Aragues-19ACC}}]
\label{lemma_properties}
Consider $n$ robots executing algorithm~\ref{algo_main}. The method satisfies the following facts:
\begin{itemize}
\item $(i)$ $\mathop{\sum}_{i=1}^n o_i(t)$ remains constant for all $t$.
\item $(ii)$ The regions associated to each robot are disjoint, with the only common point being the boundary.
\item $(iii)$ In the discrete asynchronous behavior (Def.~\ref{def_discreteAsynchVersion}) the order of the robots is preserved.
\end{itemize}
\end{lemma}
\begin{IEEEproof}
See Appendix \ref{appendices_lemmas_weigthedConsensus}.
\end{IEEEproof}

Depending on the relative speeds of robot $i$ and $i+1$, it may be the case that, between events involving $i$, $i+1$ they exchange positions. E.g., if 
$y_i(t_e^+) > y_i(t_e)$, and $v_i>>v_{i+1}$, robot $i$ may get to $y_{i-1}(t_e)$ 
and get back to $y_i(t_e^+)$ before robot $i+1$ has reached $y_i(t_e^+)$. This is temporary: robot $i$ will stop at $y_i(t_e^+)$, but robot $i+1$ will continue to $y_{i+1}(t_e)\geq y_i(t_e^+)$. Thus, in the discrete asynchronous behavior (Def.~\ref{def_discreteAsynchVersion}), the order of the robots is preserved, and robots do not need to e.g., exchange identifiers.

Now, we discuss the joint connectivity of the network. We prove that each robot $i=1,\dots,n$ meets its neighbors $i-1$ and $i+1$ after some bounded amount of time. 

\begin{proposition}[Joint connectivity{\cite[Prop. 5.2]{Aragues-19ACC}}]
\label{prop_jointconn}
Algorithm~\eqref{algo_main} under Assumptions $(A1)$, $(A2)$, gives rise 
 to a network 
 in which the set of communication graphs that occur infinitely often are jointly connected.
\end{proposition}
\begin{IEEEproof}
See Appendix \ref{appendices_prop_jointconn}.
\end{IEEEproof}

The proof of Prop. \ref{prop_jointconn} uses Assumption  \ref{ass_method}($A2$) and Lemma  \ref{lemma_properties}($ii$) in order to ensure that, during all the executions of the algorithm, at least two robots will have different orientations. This property is key in order to prove that the algorithm does not exhibit blocking (robots never get blocked waiting at different boundaries).

We are ready to prove Theorem ~\ref{th_common_traversing_times} (see Appendix \ref{appendices_ths}).

The analysis of the Discovery and Catch phase (Algorithm~\ref{algo_discovery}) that takes place during the first time instants is omitted for clarity. Note that it could be done in a similar way to the previous analysis of Algorithm~\ref{algo_main}.

\section{Performance for Interlaced Orientations}
\label{sec_interlaced}

In this section and the next one, we present the tools to prove Th.~\ref{th_performance_balanced} and \ref{th_performance_unbalanced} regarding the performance of the method. All along the section, we will assume robots have already run enough iterations of the algorithm (Algs. \ref{algo_discovery}, \ref{algo_main}  
in Sec.~\ref{sec_method}) and they already work with $y_i^\star$, $t_\star$, $d_i^\star$ (Th.~\ref{th_common_traversing_times}) for all $i$. We first present a tool for analyzing the method using an equivalent discrete synchronous version based on the concept of \emph{rounds}. Then, we introduce the concept of balanced and unbalanced \emph{interlaced}  configurations, and we discuss the implications regarding the performance of the algorithm  (Theorems \ref{th_performance_balanced} and \ref{th_performance_unbalanced}). Later, in Sec.~\ref{sec_interlacement_convergence}, we prove the  convergence of the method to these interlaced configurations. 

\subsection{Round--based method}

\begin{assumption}
\label{ass_commonTraversingTimes}

We assume robots have run the method in Section~\ref{sec_method} (Algs. \ref{algo_discovery}, \ref{algo_main}) for a time $t_0$ that we  will call \emph{initial time}. We assume $t_0$ is large enough, so that the traversing times have already converged to the common traversing time, $e_i(t_0)=e_i^\star=t_\star$, with $t_\star$ as in eq.~\eqref{eq_t_star_radii}, and equivalently $y_i(t_0)=y_i^\star$, $d_i(t_0)=d_i^\star$. We impose that the initial time $t_0$ is not an event time. 
\end{assumption}

\begin{remark}
Note that the convergence to the common traversing time $t_\star$ as in eq.~\eqref{eq_t_star_radii} is asymptotic (Theorem~\ref{th_common_traversing_times}) instead of  finite time. The difference between $e_i(t)$ and $t_\star$ can be anyway quite small by considering the time $t_0$ large enough. Thus, the performance discussed in this section and the next one will be in practice affected by some small perturbations, although in all our simulations we have observed almost no differences for $t_0$ large enough.
\end{remark}

\begin{definition}[Initial states and arrival times]
\label{def_study_initialization}
Under Assumption~\ref{ass_commonTraversingTimes}, at $t_0$ each robot $i\in\{1,\dots,n\}$ has an \emph{initial state} given by its position $p_i(t_0)$, orientation $o_i(t_0)$,  and activity $a_i(t_0)$.  We let $t_i^{e}$ be the time at which each robot $i$  arrives from its \emph{initial} position $p_i(t_0)$ to one of its boundaries ($t_i^{e}=t_0$ if robot $i$ is waiting at a boundary  $a_i(t_0)=0$ at the initial time $t_0$, and otherwise it is the time to arrive to boundary $y_{i-1}^\star$ if $o_i(t_0) < 0$ and  $a_i(t_0)=1$, or to boundary $y_{i}^\star$ if $o_i(t_0) > 0$ and $a_i(t_0)=1$), i.e.,
\begin{align*}
t_i^{e} &= \left\{ \begin{array}{ll} 
t_0+\frac{y_i^\star-r_i-p_i(t_0)}{v_i} &\mathrm{if~} o_i(t_0)>0 \mathrm{~and~} a_i(t_0)=1,\\
t_0+\frac{p_i(t_0)- y_{i-1}^\star +r_i}{v_i} &\mathrm{if~} o_i(t_0)<0 \mathrm{~and~} a_i(t_0)=1,\\
t_0 &\mathrm{if~} a_i(t_0)=0.\end{array}        \right.
\end{align*}
Note that $t_i^{e}\in [t_0, t_0+t_\star)$. 
\hfill\IEEEQEDhere
\end{definition}

Now we define the concept of \emph{round} and present the equivalent discrete--time synchronous version of the method.

\begin{definition}[Round $k$]
\label{def_round}
The $k-$th round, with $k=0,1,2,\dots$ is the following time interval with duration $t_\star$:
\begin{align}
[t_0 + k t_\star,  t_0 + (k+1) t_\star),
\label{eq_round}
\end{align}
with the initial time $t_0$ as in Assumption~\ref{ass_commonTraversingTimes}.\hfill \IEEEQEDhere
\end{definition}

\begin{definition}[Discrete Synchronous Behavior]
\label{def_behaviorDiscreteSynch}
The discrete synchronous version of the algorithm in Section~\ref{sec_method} 
(Alg.~\ref{algo_main}) under the conditions in 
Assumption~\ref{ass_commonTraversingTimes} and the initial states in Def.~\ref{def_study_initialization}, updates states and schedules events at each \emph{round} $k$ (Def.~\ref{def_round}). 
Instead of the time $t$ or the event time $t_e$, in the discrete synchronous  behavior, variables include the round $k$ associated to the time interval \eqref{eq_round}.
 For each robot $i$, the algorithm keeps track of its position $p_i(k)\in\{y^*_{i-1}+r_i, y^*_{i}-r_i\}$, orientation $o_i(k)\in\{-1,+1\}$, 
and the event times  $t^e_i(k)$ at which robot $i$ arrives to its boundaries 
 (it is not necessary to keep track of $a_i(k)$). 
At round $k=0$, these states are initialized for $i=1,\dots,n$ with
\begin{align}
t^e_i(0)&=t^e_i, &p_i(0)&=p_i(t^e_i), &o_i(0)&=o_i(t_0),
\label{eq_dis_synch_init}
\end{align}
where $t^e_i$ is given in Def. \ref{def_study_initialization}. 
 Variable $t^e_i(k)$ represents the latest time of arrival to a boundary. If an arrival takes place during the current round $k$ (eq.~\eqref{eq_round}), then $t^e_i(k) \in [t_0 + k t_\star,  t_0 + (k+1) t_\star)$.  Otherwise, $t^e_i(k)$ represents a time that belongs to a previous round, $t^e_i(k)<t_0 + k t_\star$. Variable 
$o_i(k)$ represents the orientation of robot $i$ at the beginning of round $k$. Variable $p_i(k)$ represents the position of robot $i$ at the time $t^e_i(k)$ of its latest arrival to a boundary, i.e., $p_i(k)=p_i(t^e_i(k))$. If the arrival takes place during the current round, then the robot position $p_i(t)$ takes the value $p_i(k)$ during the current round, at time $t=t^e_i(k)$. If the arrival took place in a previous round, then the robot position equals $p_i(k)$ at the beginning of the round.

At round $k$, with $k=0,1,2,\dots$, there is a meeting between each pair $(i,i+1)$ of robots satisfying 
\begin{align}
p_i(k)-r_i&=p_{i+1}(k)+r_i, \mathrm{~equivalently,} \notag\\
o_i(k)&=+1, \mathrm{~and~} o_{i+1}(k)=-1.
\label{eq_dis_synch_meet_k_o}
\end{align}
The time  at which the meeting event takes place within round $k$ is $\max\{t_i^e(k), t_{i+1}^e(k)\}$. Note that when \eqref{eq_dis_synch_meet_k_o} is satisfied, the meeting takes place during the round $k$, in particular at time $\max\{t_i^e(k), t_{i+1}^e(k)\}$.
Note also that within a round $k$, there may be several different meeting events.

The states of robots $i$ not involved in meetings remain unchanged during the next round $k+1$, i.e., $p_i(k+1)=p_i(k)$, $o_i(k+1)=o_i(k)$, $t_i^e(k+1)=t_i^e(k)$. For all the $(i,i+1)$ robots involved in meetings, the states are  updated to show their arrivals to the boundary during the next round $k+1$:
\begin{align}
p_i(k+1)&=y_{i-1}^\star + r_i, &&p_{i+1}(k+1)=y_{i+1}^\star - r_{i+1},\notag\\
o_i(k+1)&=-1,&& o_{i+1}(k+1)=+1,\label{eq_dis_synch_upd}\\
t_{i+1}^e(k+1)&=t_i^e(k+1)&&= \max\{t_i^e(k), t_{i+1}^e(k)\} + t_\star.\notag
\end{align}
Note that $t_i^e(k+1)= \max\{t_i^e(k), t_{i+1}^e(k)\} + t_\star$ is the event time at which robots will arrive to the opposite boundary, since $\max\{t_i^e(k), t_{i+1}^e(k)\}$ is the time at which the meeting took place during round $k$, and $t_\star$ is the common traversing time needed by robots to arrive to the opposite boundary after the meeting.
\hfill\IEEEQEDhere
\end{definition}

\begin{lemma}[Rounds]
\label{lemma_rounds}
 Consider the behavior of the discrete asynchronous version (Def.~\ref{def_discreteAsynchVersion}) of the method (Algorithm~\ref{algo_main}) under Assumptions \ref{ass_method} and \ref{ass_commonTraversingTimes}, and the discrete synchronous behavior (Def. \ref{def_behaviorDiscreteSynch}). The following properties hold:
$(i)$ If at round $k$ there is a meeting between robots $(i,i+1)$, with event time 
$t_e$ and updates given by eq. \eqref{eq_discreteAsynchVersion_event_time} then, 
during round $k+1$ there are exactly two arrivals to boundaries at times $t^e_i(k+1), t^e_{i+1}(k+1)$ given by \eqref{eq_dis_synch_upd}. 
During round $k$, there are no additional events (arrivals or meetings) involving $i$ or $i+1$. 

$(ii)$ Equation \eqref{eq_dis_synch_upd} encodes the same robot positions and orientations as  the discrete asynchronous version (Def.~\ref{def_discreteAsynchVersion}), for the event times $t^e_i(k)$ associated to arrivals to boundaries, i.e., $p_i(k)=p_i((t^e_i(k))^+)$, $o_i(k)=o_i(t^e_i(k))$, regardless of the order in which events take place at round $k$.

$(iii)$ Condition $o_i(k)=+1$ and $o_{i+1}(k)=-1$  and condition $p_i(k)-r_i=p_{i+1}(k)+r_i$ associated to round $k$ represent the same information.
\end{lemma}
\begin{IEEEproof}
See Appendix \ref{appendices_lemmas_Rounds_interlaced}.
\end{IEEEproof}

An example can be observed in Figure~\ref{fig_interlaced_Synch_3}. Rounds are separated by vertical gray solid lines and have duration $t_\star=127.8$. During round $k=1$, robot $i=6$ has states $p_i(k)=y^\star_6 -r_6$, $o_i(k)=+1$, $t^e_i(k)=175.9=k*t_\star + 48.1$. For all the duration $t_\star=127.8$ of the round, robot $i=6$ reaches at most one of its boundaries.

\subsection{Interlaced Orientations}

Now, we use the concept of \emph{rounds} (Definition~\ref{def_round}) and the equivalent  representation of the algorithm based on the discrete synchronous version (Definition~\ref{def_behaviorDiscreteSynch}), to study the performance of the method for balanced and unbalanced orientations (Def.~\ref{def_balanced_unbalanced}). In this section, we will consider orientations which are \emph{interlaced}, which is a concept we explain next. After this, in section~\ref{sec_interlacement_convergence}, we will prove that in fact, the method interlaces the orientations.

\begin{definition}[Interlaced Orientations]
\label{def_interlaced}
Let $n_+$, $n_-$, and $n_{bal}$ be as in Def.~\ref{def_balanced_unbalanced}. The robot orientations are \emph{interlaced} at round $k$ when there exist distinct robot indexes 
\begin{align*}
i_1, i_2, \dots, i_{n_{bal}}
\end{align*}
such that, for all $j=1,\dots,n_{bal}$:
\begin{align}
o_{i_j}(k)=+1 ~\mathrm{and}~ o_{i_j + 1}(k)=-1.
\label{eq_interlaced_index}
\end{align}
Note that in case $n_+ > n_-$, the orientations of the remaining robots are positive, and if the orientations are balanced ($n_{bal}=n/2$), there are no remaining robots.
\hfill\IEEEQEDhere
\end{definition}

\begin{figure}[tbp]
\begin{center}
\begin{tabular}{cc}
\includegraphics[width=0.18\paperwidth]{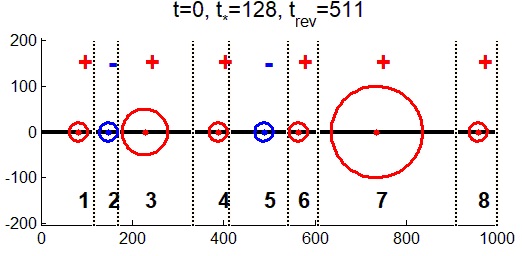}&
\includegraphics[width=0.18\paperwidth]{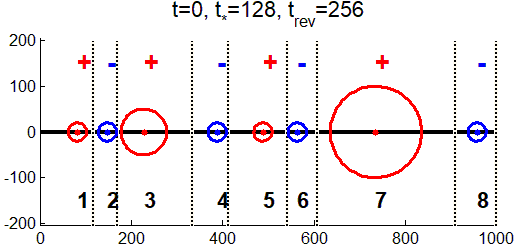}
\end{tabular}
\end{center}
\caption{Two examples of robots running the method in Section~\ref{sec_method}
 (Algs. \ref{algo_discovery},\ref{algo_main})
 with different maximum speeds and communication radii (circles around the robots), moving forward (red) and backward (blue) between their boundaries $y_i^\star$ \eqref{eq_optimal_boundary} (gray solid, in vertical) as in Asm.~\ref{ass_commonTraversingTimes}, Def.~\ref{def_study_initialization}. {\bf Left}: Unbalanced interlaced orientations (Defs.~\ref{def_balanced_unbalanced} and  \ref{def_interlaced}), with $n_{bal}=2$, and with robot indexes  $i_1=1, i_2=4$ at round $k=0$ ($o_1(0)=+1, o_2(0)=-1$, $o_4(0)=+1, o_5(0)=-1$, and the remaining orientations are positive). 
{\bf Right}: Balanced interlaced orientations (Defs.~\ref{def_balanced_unbalanced}, and \ref{def_interlaced}), with $n_{bal}=n/2$.
}
\label{fig_interlaced_Synch_1}
\end{figure}

An example of interlaced orientations can be seen in Fig.~\ref{fig_interlaced_Synch_1} {\bf Left}: $n=8$, $n_{bal}=2$, and $i_1=1, i_2=4$, so that $o_1(k_0)=+1, o_2(k_0)=-1$, and $o_4(k_0)=+1, o_5(k_0)=-1$, and the other robots have positive orientation. The orientations in Fig.~\ref{fig_interlaced_Synch_1} {\bf Right}, are  interlaced as well, and balanced in this case: $n=8$, $n_{bal}=n/2=4$, and $i_1=1, i_2=3, i_3=5, i_4=7$, so that $o_1(k_0)=o_3(k_0)=o_5(k_0)=o_7(k_0)=+1$, and $o_2(k_0)=o_4(k_0)=o_6(k_0)=o_8(k_0)=-1$.

\begin{lemma}[Interlaced Meetings]
\label{lemma_interlaced}
Assume at round $k$ the configuration is interlaced with the robot indexes in Definition~\ref{def_interlaced} being $i_1, i_2, \dots, i_{n_{bal}}$. Without loss of generality, assume $n_+ \geq n_-$. Then:
$(i)$ At round $k+1$ the configuration is interlaced, with robot indexes $i_1-1, i_2-1, \dots, i_{n_{bal}}-1$.
$(ii)$ At all the successive rounds $k' \geq k$ the configuration remains interlaced.
$(iii)$ At all the successive rounds $k' \geq k$ there are exactly $n_{bal}$ meetings.
$(iv)$ Each robot $i=1,\dots,n$ arrives to $y_i^\star$ boundary $n_{bal}$ times every $n$ rounds.
\end{lemma}
\begin{IEEEproof}
See Appendix \ref{appendices_lemmas_Rounds_interlaced}.
\end{IEEEproof}

\begin{proposition}[Unbalanced Interlaced Performance]
\label{prop_performance_unbalanced_interlaced}
Under assumption~\ref{ass_commonTraversingTimes}, a robot team with unbalanced interlaced orientations (Defs.~\ref{def_balanced_unbalanced} and ~\ref{def_interlaced}) running the algorithm in Section~\ref{sec_method} (Alg. \ref{algo_main}),  has a performance in terms of the revisiting times $t_{rev}$ (Def.~\ref{def_revisiting_time}) averaged along $n_{bal}$ meetings, given by 
\begin{align}
t_{rev}=t_\star n  / n_{bal}.
\end{align}
\end{proposition}
\begin{IEEEproof}
See Appendix \ref{appendices_props_interlaced_performance_synch}.
\end{IEEEproof}

Moreover, when the orientations are not only \emph{interlaced} but also \emph{balanced},  as the following result shows, robots synchronize to perform exactly their meetings simultaneously in the network.

\begin{proposition}[Balanced Interlaced Synchronization]
\label{prop_balanced_interlaced_clock_synch}
Assume the conditions in Assumption~\ref{ass_commonTraversingTimes}, Def.~\ref{def_study_initialization} hold and there is a round $k_0$ at which the robots achieve a balanced interlaced configuration (Defs.~\ref{def_balanced_unbalanced}, ~\ref{def_interlaced}). Let $t'_0$ be the initial time of round $k_0$, i.e., $t'_0=t_0 + k_0*t_\star$, and $t^e_1(k_0),\dots,t^e_n(k_0)$ be the time of arrival to boundary of each robot $i$ at round $k_0$, defined as $t_i^e$ in Def.~\ref{def_study_initialization} and \eqref{eq_dis_synch_init} using the new $t'_0$ instead of $t_0$. Then, after $n/2$ rounds, all robots $i=1,\dots,n$  synchronize their  event times, meaning that, for all rounds $k\geq k_0 + n/2$, all the $n/2$ meeting events associated to the round $k$ take place simultaneously at time
\begin{align}
t^{e}_i(k)&= (k-k_0) t_\star + \max_j\{ t^e_j(k_0) \}.
\label{eq_balanced_interlaced_synchronized_event_times}
\end{align}
After these $n/2$ rounds, from then on, the revisiting times $t_{rev}$ are exactly 
\begin{align}
t_{rev}= 2 t_\star.
\end{align}
\end{proposition}
\begin{IEEEproof}
See Appendix \ref{appendices_props_interlaced_performance_synch}.
\end{IEEEproof}

\begin{figure}[tbp]
\begin{center}
\includegraphics[width=0.4\paperwidth]{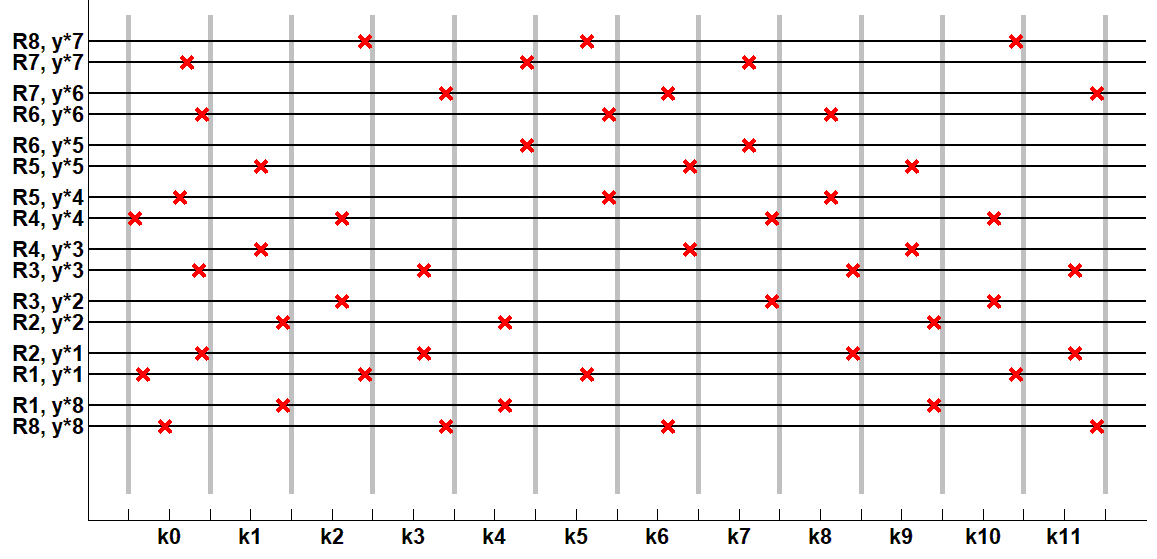}\\
\end{center}
\caption{Time of arrival (red crosses) of each robot $i$ to boundaries $y^\star_{i-1}$ and $y^\star_i$ (y--axis) along time and rounds (x--axis), for the scenario in Fig.~\ref{fig_interlaced_Synch_1} {\bf Left} (unbalanced interlaced orientations). 
 Gray solid vertical lines represent the separation between rounds $k=0,1,2,\dots$ (Def.~\ref{def_round}), where we take $k_0=0$. 
 Each robot $i$ arrives at the $y_i^\star$  boundary (red crosses on lines $R_i, y^\star_i$) but for robots $i=2$ and $i=5$, which arrive at the  $y_{i-1}^\star$ boundary, (red crosses on lines $R_2, y_1^\star$ and $R_5,y_4^\star$). 
 At round $k=1$, orientations are unbalanced interlaced, with indexes $i_1-1=1, i_2-1=4$ (red crosses on lines $R_1, y_n^\star$ and $R_4,y_3^\star$), and so on. At each round there are $n_{bal}=2$ meetings involving $2 n_{bal} = 4$ robots (4 red crosses at each round $k$) as in Lemma~\ref{lemma_interlaced}. Every $n$ rounds, e.g., $1\leq k \leq 7$, or $2\leq k \leq 8$, each robot $i$ arrives $n_{bal}$ times to boundary $y_i^\star$ (red crosses).
}
\label{fig_interlaced_Synch_2}
\end{figure}

\begin{figure}[tbp]
\begin{center}
\includegraphics[width=0.4\paperwidth]{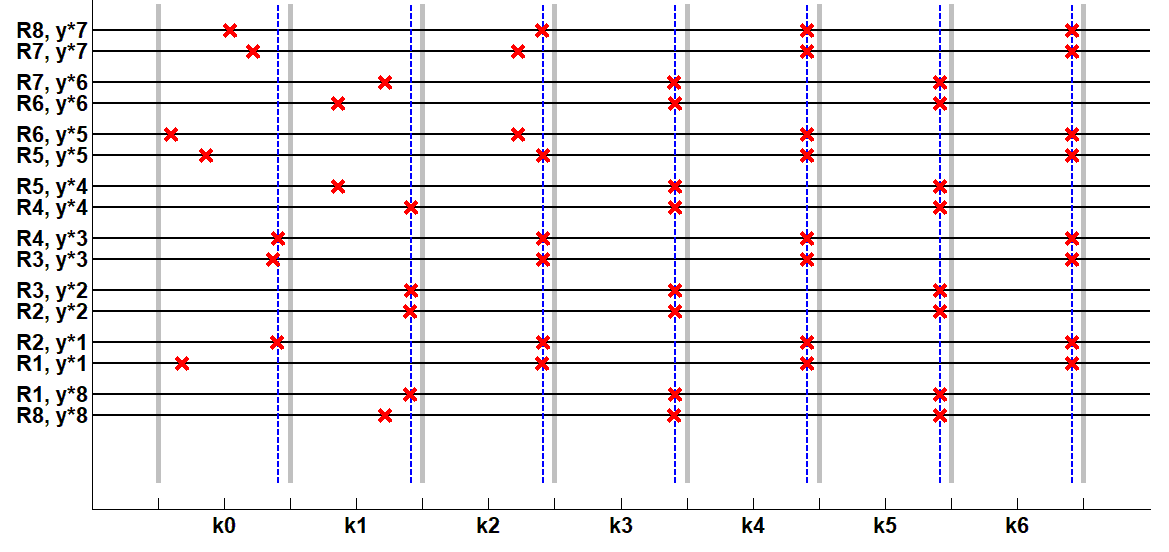}
\end{center}
\caption{
Time of arrival (red crosses) of each robot $i$ to boundaries $y^\star_{i-1}$ and $y^\star_i$ (y--axis) along time and rounds (x--axis), for the scenario in Fig.~\ref{fig_interlaced_Synch_1} {\bf Right} (balanced interlaced orientations). 
Gray solid vertical lines represent the separation between rounds $k=0,1,2,\dots$ (Def.~\ref{def_round}), where we take $k_0=0$. At every round $k$ there are $n_{bal}=n/2$ meetings involving $2 n_{bal} = n$ robots (red crosses). After $n/2=4$  rounds, robots synchronize their arrival times (red crosses) according to eq.~\eqref{eq_balanced_interlaced_synchronized_event_times} (blue dashed vertical lines) as in Prop.~\ref{prop_balanced_interlaced_clock_synch}.
}
\label{fig_interlaced_Synch_3}
\end{figure}

Figures~\ref{fig_interlaced_Synch_1},~\ref{fig_interlaced_Synch_2},~and \ref{fig_interlaced_Synch_3} show an example of the previous properties.

\section{Convergence to Interlaced Orientations}
\label{sec_interlacement_convergence}

In the previous section, we have characterized the system performance, assuming that the robot orientations were interlaced (Def.~\ref{def_interlaced}). In this section we prove that in fact, the method in Sec.~\ref{sec_method} (Algs. \ref{algo_discovery}, \ref{algo_main}) makes the orientations become interlaced in a finite number of rounds. Then, as stated by Lemma~\ref{lemma_interlaced}, the interlaced configuration will be kept for all subsequent rounds, giving the performance in Prop.~\ref{prop_performance_unbalanced_interlaced}, \ref{prop_balanced_interlaced_clock_synch}. As in the previous section, here we assume the  conditions in Assumption~\ref{ass_commonTraversingTimes} and Def.~\ref{def_study_initialization} hold, and we analyze the algorithm using the Discrete Synchronous version of the method (Def.~\ref{def_behaviorDiscreteSynch}) and the concept of round $k$ (Def.~\ref{def_round}). 

Since the interlaced property (Def.~\ref{def_interlaced}) depends exclusively on the robot orientations, in this section we focus on the evolution of the  orientations along different rounds $o_i(k)$. We represent with $+$ and $-$ positive and negative orientations ($o_i(k)>0$, $o_i(k)<0$), and study how they evolve under the meeting events \eqref{eq_dis_synch_upd}.

\begin{figure*}[tbp]
\begin{center}
\begin{tabular}{cc}
\begin{tabular}{cc}
{\footnotesize $k=0$}&\includegraphics[width=0.3\paperwidth]{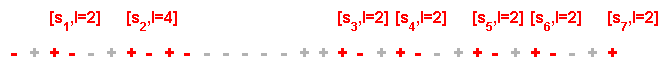}\\
{\footnotesize $k=1$}&\includegraphics[width=0.3\paperwidth]{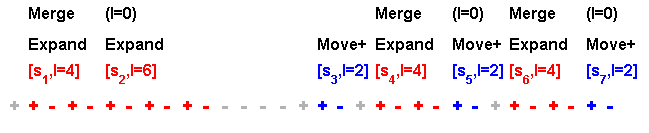}\\
{\footnotesize $k=2$}&\includegraphics[width=0.3\paperwidth]{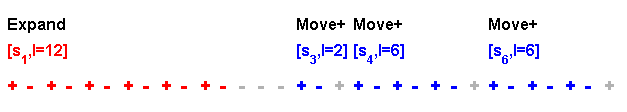}\\
{\footnotesize $k=3$}&\includegraphics[width=0.3\paperwidth]{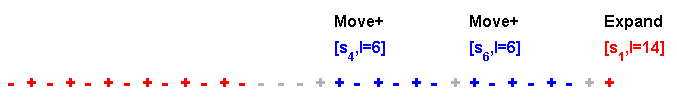}\\
{\footnotesize $k=4$}&\includegraphics[width=0.3\paperwidth]{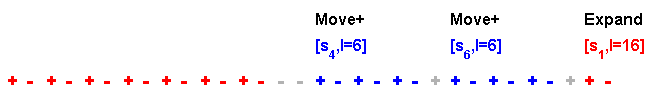}
\end{tabular} &
\begin{tabular}{ll}
{\footnotesize $k=0$}&\includegraphics[height=0.0258\paperwidth]{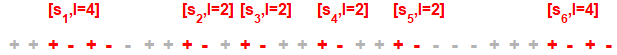}\\
{\footnotesize $k=1$}&\includegraphics[height=0.04\paperwidth]{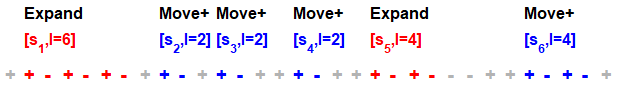}\\
{\footnotesize $k=2$}&\includegraphics[height=0.04\paperwidth]{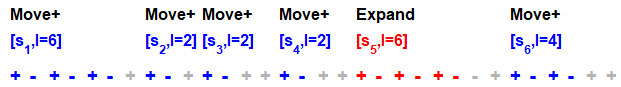}\\
{\footnotesize $k=3$}&\includegraphics[height=0.052\paperwidth]{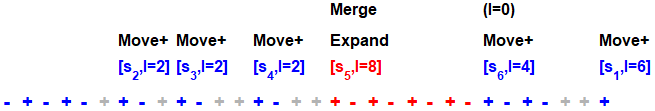}\\
{\footnotesize $k=4$}&\includegraphics[height=0.04\paperwidth]{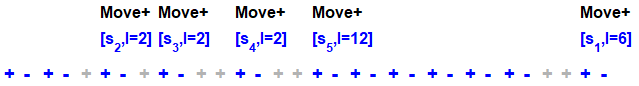}
\end{tabular}\\
\includegraphics[height=0.15\paperwidth]{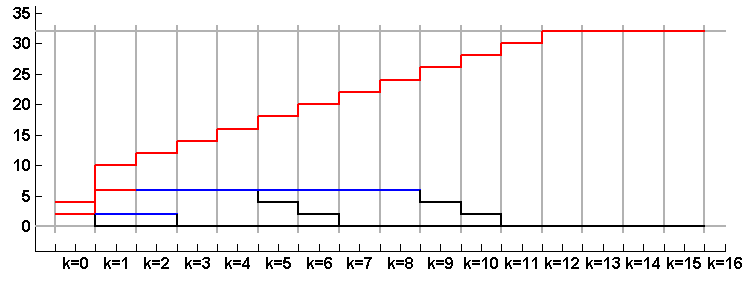}&
\includegraphics[height=0.15\paperwidth]{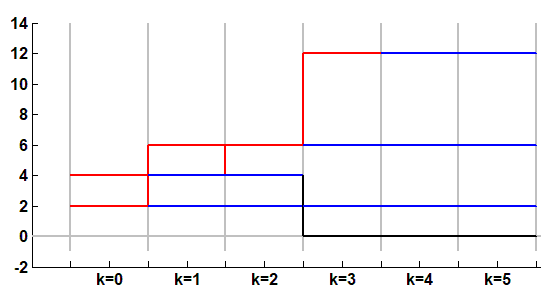}
\end{tabular}
\end{center}
\caption{Interlacing for two examples. 32 robots run the method in Section~\ref{sec_method} (Alg. \ref{algo_main}) with different maximum speeds and communication radii, under the conditions in Assumption~\ref{ass_commonTraversingTimes} and Def.~\ref{def_study_initialization}. In one example  ({\bf Left column}), robot orientations are balanced (Def.~\ref{def_balanced_unbalanced}), and in the other example ({\bf Right column}) orientations are unbalanced, with $n_+=20$, $n_-=12$, $n_{bal}=12$. 
{\bf Top} We show the evolution of the sequences and letters associated to the robot orientations along some rounds, where $+$ means $o_i(k)>0$ and $-$ means  $o_i(k)<0$. At each round $k$, we display the rule that was applied to transition from $k-1$ to $k$.
{\bf Bottom}: Evolution along the rounds  $k=0,1,2,\dots$ of the length of every sequence. 
Note that, initially ($k=0$), orientations are not interlaced (Def.~\ref{def_interlaced}).}
\label{fig_interlacing}
\end{figure*}

\begin{definition}[Orientation words and sequences]
\label{def_orienation_words_sequences}
Given a specific round $k$, we represent the robot orientations with a word $w(k)$ of length $n$  consisting of $+$ and $-$ characters. We let $w(k)_{j1}$ be the element placed at position $j1$ in $w(k)$, and $w(k)_{j1:j2}$ be the elements in the orientation word $w(k)$ between positions $j1$ and $j2$. We use $[seq,l]$ (\emph{sequence}) to refer to $l$ consecutive elements, which start with a $+$ element, which has the same number of $+$ and $-$ elements, and which is interlaced:
\begin{align}
[seq,l] &= + - \dots + -.
\end{align}
Note that the length $l$ associated to the sequence is an even number ($l=2, 4, \dots$).  We use the term \emph{letter} to refer to the $+$, $-$ characters in the word $w(k)$ which do not belong to any  sequence. 
There may be several sequences $[seq,l], [seq',l'],[seq'',l''],\dots$ in $w(k)$ with different lengths,  and sequences are separated by one or more letters. 
\begin{align*}
w(k)&= -\dots - + \dots + [seq,l]-\dots - + \dots + [seq',l'] \dots
\end{align*}
\hfill\IEEEQEDhere
\end{definition}
Some examples of orientation words and sequences are given in Figure~\ref{fig_interlacing}.

We first discuss the evolution of the sequences after each round. We show that this evolution depends on the  letters around it, being represented by different rules. After this, we discuss the evolution of sequences under these rules, and the effect on the word. We finally discuss the convergence of $w(k)$ to interlaced configurations. 

\begin{lemma}[Sequence evolution]
\label{lemma_sequence_evolution_rules}
Consider the orientation word $w(k)$ at round $k$. Every sequence $[seq, l]$ placed in $w(k)$ in positions $w(k)_{j1:j1+l-1}$ evolves between rounds $k$ and $k+1$ depending on the letters  around it at positions 
$w(k)_{j1-1}$ and $w(k)_{j1+l}$, according to the following rules:
\begin{itemize}
\item \emph{Move+} rule: The sequence moves one position to the left, consuming the $+$ letter to its left, and providing a $+$ letter to its right:
\begin{align*}
\mathrm{if~} w(k)_{j1-1:j1+l}=   & &+ [seq, l] &+  \\
\mathrm{then~} w(k+1)_{j1-1:j1+l}=  && [seq, l] +  &(+)
\end{align*}
\item \emph{Move-} rule: The sequence moves one position to the right, 
 consuming the $-$ letter to its right, and providing a $-$ letter to its left:
\begin{align*}
\mathrm{if~} w(k)_{j1-1:j1+l}=   &&& ~-~[seq, l] -  \\
\mathrm{then~} w(k+1)_{j1-1:j1+l}=  &&&(-)-  [seq, l] 
\end{align*}
\item \emph{Expand} rule: The sequence \emph{Expands}, increasing its length,  and consuming both the $+$ letter to its left and the $-$ letter to its right:
\begin{align*}
\mathrm{if~} w(k)_{j1-1:j1+l}=    &&+ [seq, l] -  \\
\mathrm{then~} w(k+1)_{j1-1:j1+l}=  &&[seq, l+2]
\end{align*}
\item \emph{Reduce} rule: The sequence \emph{Reduces}, decreasing its length,  and providing both a $+$ letter to its right and a $-$ letter to its left:
\begin{align*}
\mathrm{if~} w(k)_{j1-1:j1+l}=    &&& -~ [~seq, l+2~] ~+  \\
\mathrm{then~} w(k+1)_{j1-1:j1+l}=  &&&(-) -[seq, l]+(+)
\end{align*}
The remaining elements in $w(k)$ not affected by these rules (letters not surrounding sequences), remain the same in $w(k+1)$. The elements that appear between parenthesis $(+)$, $(-)$, take in fact these values in $w(k+1)$ but, due to possible interactions between  sequences (other sequences may consume these elements), they may not be letters but be part of a sequence.
\item \emph{Merge} rule:
In addition, after \emph{Move+}, \emph{Move-}, or \emph{Expand} rules, the sequence may \emph{Merge} with sequences in the left, the right, or both. The \emph{Merge} rule, if any, takes place at round $k+1$, and it is applied as many times as necessary to ensure sequences are correctly organized. 
\begin{align*}
\mathrm{if~} w(k+1)_{j1:j1+l+l'-l}= &&[seq, l] [seq', l']  \\
\mathrm{then~} w(k+1)_{j1:j1+l+l'-l}=  &&[~seq,~ l + l'~]
\end{align*}
\end{itemize}
\end{lemma}
\begin{IEEEproof}
See Appendix \ref{appendices_lemmas_sequences_basic}.
\end{IEEEproof}

\begin{lemma}[Sequence properties]
\label{lemma_sequence_properties}
Consider the orientation word $w(k)$ and the sequences and letters acting according to Lemma~\ref{lemma_sequence_evolution_rules}. The following properties hold:
\begin{itemize}
\item ($i$): Sequences move at most one position to the left and/or to the right per round.
\item ($ii$): No new sequences can be created.
\end{itemize}
\end{lemma}
\begin{IEEEproof}
See Appendix \ref{appendices_lemmas_sequences_basic}.
\end{IEEEproof}

Lemma~\ref{lemma_sequence_evolution_rules} states the sequence evolution rules between  consecutive rounds $k$ and $k+1$. Now, we study how the sequences $[seq, l]$ (Def.~\ref{def_orienation_words_sequences}) evolve \emph{along several rounds}. Recall that, without loss of generality, we assume $n_+ \geq n_-$. As we will show, several sequences act in a collaborative way: they move all the $+$ letters they find to their right, or the $-$ letters to their left, to aid the remaining  sequences. In addition, several sequences   disintegrate, placing their own $+$ and $-$ letters respectively to their right and their left, so that they can be used by the other sequences.

\begin{lemma}\emph{(Sequences that Reduce):}
\label{lemma_collaborative_sequences_reduce}
As long as a sequence $[seq,l]$ experiences the \emph{Reduce} rule for the first time, it keeps on running the \emph{Reduce} rule until it disappears after $l/2$ rounds. At each of these rounds,  this sequence provides a $+$ letter through its right, and a $-$ letter through its left. 
\end{lemma}
\begin{IEEEproof}
See Appendix \ref{appendices_lemmas_sequences_basic}.
\end{IEEEproof}

\begin{lemma}\emph{(Sequences that Move-):}
\label{lemma_collaborative_sequences_move_neg}
As long as a sequence $[seq,l]$ experiences the \emph{Move-} rule for the first time, it keeps on running the \emph{Move-} rule while there are $-$ letters to the right. It eventually disappears (\emph{Reduce}), placing at every round one $+$ letter to its right, and one $-$  letter to its left. If during this process, it \emph{Merges} with another sequence $[seq',l']$, both sequences eventually disappear (\emph{Reduce}).
\end{lemma}
\begin{IEEEproof}
See Appendix \ref{appendices_lemma_move_neg}.
\end{IEEEproof}

\begin{lemma}\emph{(Sequences that Move+):}
\label{lemma_collaborative_sequences_move_pos}
As long as a sequence $[seq,l]$ experiences the \emph{Move+} rule for the  first time, it keeps on running the \emph{Move+} rule while there are $+$ letters to the left. 
($i$) When  orientations are balanced as in Def.~\ref{def_balanced_unbalanced} ($n_{bal}=n/2$, i.e., $n_+ = n_-$), the sequence eventually disappears (\emph{Reduce}, Lemma~\ref{lemma_collaborative_sequences_reduce}) providing $+$ letters to its right, and $-$ letters to its left. If during this process, it \emph{Merges} with other sequence $[seq',l']$, both sequences eventually disappear (\emph{Reduce}).
($ii$) When orientations are unbalanced ($n_{bal}<n/2$) the sequence may either behave as in ($i$), or it may keep on running the \emph{Move+} rule for ever. If during this process, it \emph{Merges} with other sequence $[seq',l']$, both  sequences run the \emph{Move+} rule for ever. This represents interlaced orientations (Def.~\ref{def_interlaced}) and meetings that take place according to Lemma~\ref{lemma_interlaced} and its proof.
\end{lemma}
\begin{IEEEproof}
See Appendix \ref{appendices_lemma_move_pos}.
\end{IEEEproof}

Now, we prove that the orientation word $w(k)$ evolves until an interlaced configuration (Def.~\ref{def_interlaced}) is reached. This is achieved when $w(k)$ is exclusively composed of $+$ letters, and sequences with lengths summing up to $2 n_{bal}$ (recall we assume $n_+\geq n_-$). In our analysis, we will consider the evolution of the sequences (Lemmas~\ref{lemma_collaborative_sequences_reduce}, ~\ref{lemma_collaborative_sequences_move_neg}, ~\ref{lemma_collaborative_sequences_move_pos}) and we will show that, from the initial sequences, at least one of them has the property of being expansive, meaning that at every round, it consumes a $+$ and a $-$ letter, and its length is  increased by 2, until there are no $-$ letters in the word $w(k)$, i.e., the lengths of the existing sequences sum up to $2 n_{bal}$.

\begin{proposition}[Always Expand sequence]
\label{prop_always_expand_move}
Assume the conditions in Assumption~\ref{ass_commonTraversingTimes}, Def.~\ref{def_study_initialization} hold for some round $k_0$ that we take here as $k_0=0$.  Consider the initial orientation word $w(0)$  associated to the robot orientations (Def.~\ref{def_orienation_words_sequences}). Then:
 ($i$) From the initial set of sequences $[seq, l]$ in $w(0)$, at least one sequence experiences the \emph{Expand} rule during all the rounds, until all sequences in $w(k)$ have lengths summing up to $2 n_{bal}$.  
 ($ii$) The number of rounds required to achieve this interlaced configuration is smaller than $n_{bal}$.
($iii$) When the orientations are balanced ($n=2 n_{bal}$), after $(i)$ is achieved, there is   a single sequence with length $n$.
($iv$) When the orientations are unbalanced with $n_+>n_-$, after $(i)$ is achieved, then all the sequences run the \emph{Move+} rule for ever. 
($v$) The orientations become \emph{interlaced} (Def.~\ref{def_interlaced}) in a finite number of rounds.
\end{proposition}
\begin{IEEEproof}
See Appendix \ref{appendices_prop_always_expand_move}.
\end{IEEEproof}

Figure~\ref{fig_interlacing} shows some examples of the properties in 
Lemmas~\ref{lemma_collaborative_sequences_reduce}, \ref{lemma_collaborative_sequences_move_neg}, \ref{lemma_collaborative_sequences_move_pos}, and  Prop.~\ref{prop_always_expand_move}.  

In the {\bf Left column}, robot orientations are balanced (Def.~\ref{def_balanced_unbalanced}). Initially  ($k=0$), there are 7 sequences (in red) and several letters, so that the orientations are not interlaced (Def.~\ref{def_interlaced}). From $k=0$ to $k=1$, four sequences \emph{Expand} (red), and their length increases by two. Also, six sequences \emph{Merge} into three longer sequences (equivalently, three sequences disappear (in black)). During rounds $k=0,1,...$, some sequences \emph{Move+} or \emph{Move-}, keeping their lengths unchanged (in blue), and eventually \emph{Reducing} (in black), making their lengths drop to zero. Only one sequence always \emph{Expands} (always in red), and it reaches finally a length equal to $n$. The process takes less than $n/2$ rounds, giving rise to  interlaced orientations (Def.~\ref{def_interlaced}).

In the {\bf Right column}, orientations are unbalanced, with $n_+=20$, $n_-=12$, $n_{bal}=12$ (Def.~\ref{def_balanced_unbalanced}). Initially ($k=0$), there are 6 sequences (in red), with lengths summing up to $16 < 2 n_{bal}=24$, so that the orientations are not interlaced (Def.~\ref{def_interlaced}). Sequences $s_1$ and $s_5$ \emph{Expand} (red), and their length increases by two at these rounds. The remaining Sequences experience the \emph{Move+} rule (in blue), keeping their lengths unchanged. After few rounds (from $k=2$ to $k=3$), only sequence $s_5$ \emph{Expands}. Also, sequence $s_5$ \emph{Merges} with $s_6$, making $s_6$ disappear (its length falls to 0). The sequence $s_5$ that always runs the \emph{Expand} rule, has now length $8+4=12$, and the sum of its own length and of the other sequences  (2 + 2 + 2 + 12 + 6=24) becomes equal to $2 n_{bal} = 24$. The process takes less than  $n_{bal}$ rounds. From then on, the five sequences always experience the \emph{Move+} rule, with the orientations interlaced (Def.~\ref{def_interlaced}).

We are finally ready to present the proofs of the Theorems~\ref{th_performance_balanced} and \ref{th_performance_unbalanced} (see Appendix \ref{appendices_ths}).

\section{Simulations}
\label{sec_experiments}

\subsection{Simulations in the $1D$ cycle graph}
\label{sec_simulations}

\begin{figure}[tbp]
\begin{center}
\begin{tabular}{cc}
\includegraphics[width=0.18\paperwidth]{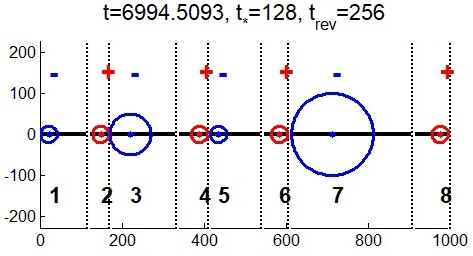}&
\includegraphics[width=0.18\paperwidth]{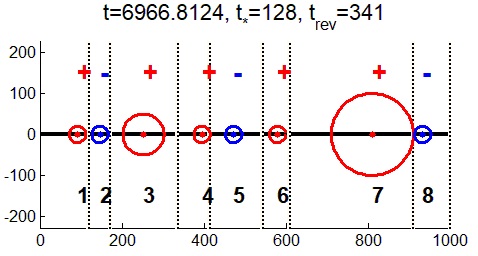}\\
\includegraphics[width=0.18\paperwidth]{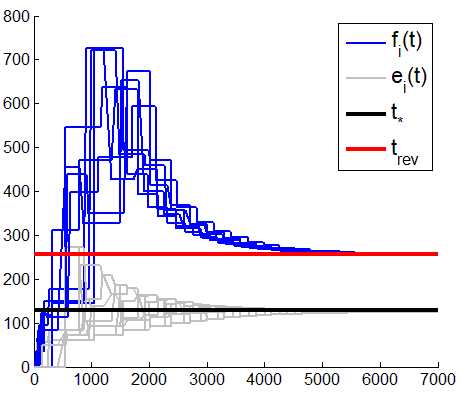}&
\includegraphics[width=0.18\paperwidth]{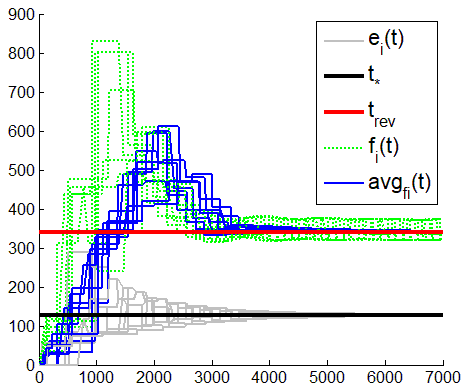}
\end{tabular}
\end{center}
\caption{Robots running Algorithms \ref{algo_discovery}, \ref{algo_main}  
in Section~\ref{sec_method} with different maximum speeds and communication radii (circles around the robots). 
 They move forward (red) and backward (blue) at their maximum speeds between their  boundaries $y_i(t)$ (black dashed, in vertical), which converge to the  boundaries $y_i^\star$ \eqref{eq_optimal_boundary}(gray solid, in vertical) associated to the common traversing time $t_\star$ \eqref{eq_t_star_radii}.  {\bf Left Top:} Final configuration for balanced orientations (Def.~\ref{def_balanced_unbalanced}). {\bf Left Bottom:}  Evolution of $e_i(t)$ ~\eqref{eq_e_def} (gray) and $f_i(t)$ (Def.~\ref{def_revisiting_time}) (blue) compared  to $t_\star$ (black) and $t_{\mathrm{rev}}$ (red) for balanced orientations. {\bf Right Top:} Final configuration for unbalanced orientations, with $n_{bal}=3$. 
{\bf Right Bottom:}  Evolution of $e_i(t)$ (gray), $f_i(t)$ (green dashed), and $f_i(t)$ averaged along $n_{bal}$ meetings (blue solid), compared  to $t_\star$ (black) and $t_{\mathrm{rev}}$ (red).
}
\label{fig_different_speeds_radii_autosynch}
\end{figure}

\begin{figure}[tbp]
\includegraphics[width=0.42\paperwidth]{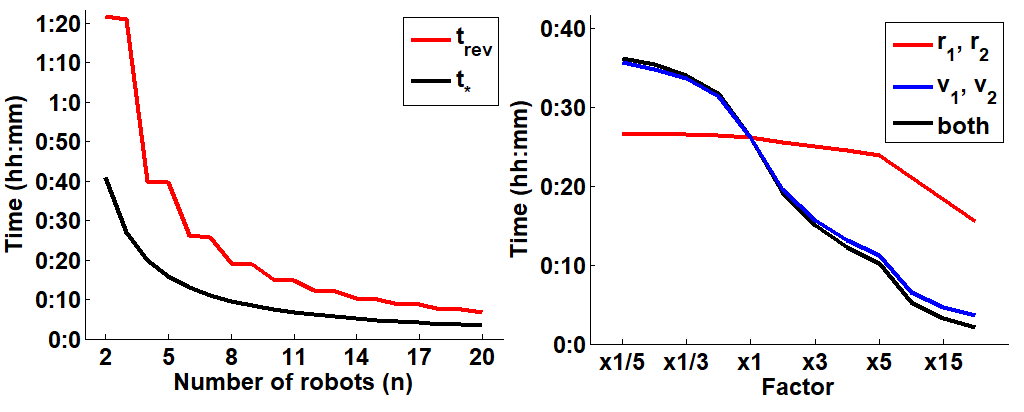}
\caption{Results of running several instances of Algorithms \ref{algo_discovery}, \ref{algo_main} in Section~\ref{sec_method}, using in each case different values of 
the number of robots $n$, the maximum speed $v_i$ and radius $r_i$, in order to analyze the effects of these parameters on the revisiting time $t_{\mathrm{rev}}$ (Th.~\ref{th_performance_balanced}, Th.~\ref{th_performance_unbalanced}). Robots run Algorithms \ref{algo_discovery}, \ref{algo_main} in Section~\ref{sec_method}.
{\bf Left:} Using a different number of robots $n$. 
{\bf Right:} Using a different maximum speed and radius for robots $i=1,2$. 
}
\label{fig_varying_n_factor_vr}
\end{figure}

Figure \ref{fig_different_speeds_radii_autosynch} shows a simulation with $n=8$  robots traversing a $1D$ cycle graph with length $L=1000$ meters. The black line between position 0 and $L$ represents the position of robots in the cycle graph (Fig.~\ref{fig_tree}). 
Robots have maximum speeds $\{v_1,\dots,v_n\}= \{0.6,0.1,0.5,0.3,0.7,0.2,0.8,0.4 \} m/s $, and communication radii equal to $20m$  apart from $r_3=50m$ and $r_7=100m$. They start randomly placed in the cycle graph, with their communication regions non overlapping. {\bf Left}: the initial orientations are balanced ($o_1(0),\dots, o_4(0)=-1$, and $o_5(0),\dots,o_8(0)=+1$). 
 Asymptotically, robots reach a configuration where their regions have common traversing times $t_\star$ which, in this balance configuration, equals $t_{\mathrm{rev}}/2$ (Theorems~\ref{th_common_traversing_times}, \ref{th_performance_balanced}). 
 Recall from Sec. \ref{sec_problem_description} that $t_\star$ and  $t_{\mathrm{rev}}=2 t_\star$ is the performance that would be achieved by a centralized system. Thus, the proposed method achieves the same performance as the centralized approach, with lighter memory costs. 
  {\bf Right}: the initial orientations are unbalanced, with $n_{bal}=3$. The  times $e_i(t)$ required to traverse the regions associated to the robots, converge to have common traversing times $t_\star$ (Theorem~\ref{th_common_traversing_times}). 
As a metric for the \emph{revisiting time} (Def.~\ref{def_revisiting_time}), in our simulations we use the \emph{inter--meeting time} $f_i(t)$, which is the time elapsed between consecutive meetings of robots $i$ and $i+1$, and which is in fact the revisiting time of the $y_i(t)$ boundary. 
 Due to the unbalanced orientations, the inter--meeting times $f_i(t)$ (green dashed), averaged along $n_{bal}$ meetings (blue solid), converge to 
$t_{\mathrm{rev}}=n t_\star / n_{bal}$ (Theorem~\ref{th_performance_unbalanced}).

We have run a set of simulations, using an increasing number of robots $n=2,\dots,20$ 
(Fig. \ref{fig_varying_n_factor_vr} {(\bf Left)}). 
All robots have maximum speed $v_i=2 m/s$ and radius $r_i=50m$, for $i=1,\dots,n$. The   number of balanced orientations equals $floor(n/2)$. The length of the cycle graph equals $10 Km$. Observe that, as the number of robots increases, the revising time $t_{\mathrm{rev}}$ (Th.~\ref{th_performance_balanced}, Th.~\ref{th_performance_unbalanced}) decreases. Fig.~\ref{fig_varying_n_factor_vr} {(\bf Right)} shows three additional sets of simulations with $6$ robots with balanced orientations collaborating to keep intermittently connected a graph cycle with length $10 Km$. 
Robots have maximum speed $v_i=2 m/s$ and radius $r_i=50m$. 
In the first set of simulations (in red), we consider different radii for robots $i=1,2$, given by $r_1=r_2=50m*factor$.
In the second set of simulations (in blue), we keep the radius constant, but we consider different speeds for robots $i=1,2$, given by  $v_1=v_2=2m/s*factor$. Finally, in the third set of simulations (in black), we let both the radii $r_1,r_2$ and speeds $v_1,v_2$ of robots $i=1,2$ to change in each case according to the associated factor. 
The $factor$ used is shown in the $x-$axis (examples $\times \frac{1}{5}$, $\times 3$, and $\times 15$). Observe that, as the robots have improved capabilities, the associated revisiting time $t_{\mathrm{rev}}$ decreases, i.e., the performance is improved. Thus, the proposed method (Algs. \ref{algo_discovery}, \ref{algo_main} in Sec.~\ref{sec_method}) makes a proper use of the available resources.

\subsection{Simulations in Gazebo/ROS}

\begin{figure}[tbp]
\begin{center}
\includegraphics[width=0.35\paperwidth]{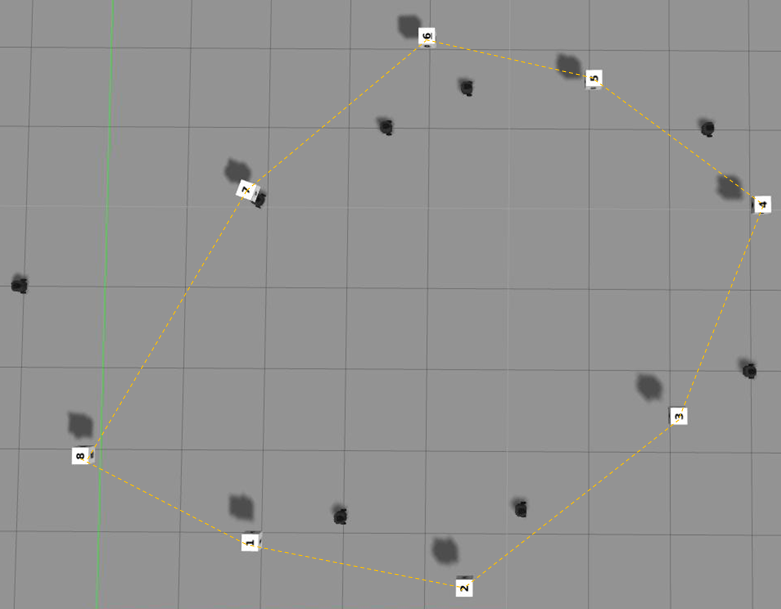}
\end{center}
\caption{Gazebo simulation. Example of scenario with 8 robots  (\emph{turtlebots}) with  balanced orientations (Def.\ref{def_balanced_unbalanced}), and with 8  tasks (white dice) on the plane that must be intermittently connected. 
}
\label{fig_Gazebo_environment}
\end{figure}

We have conducted experiments using ground robots Turtlebot 3, with differential drive motion, in a planar environment, simulated in Gabezo 7.0, under Ubuntu 16 LTS and ROS Kinetic (Fig. \ref{fig_Gazebo_environment}).

We briefly explain how the $1D$ position $p_i(t)\in[0,L]$ discussed in the paper is transformed to the $2D$ position $p^{xy}_i(t)\in\mathbb{R}^2$ in the working plane. Each robot $i$ knows the ordered set of $2D$ positions $q^{xy}_s\in\mathbb{R}^2$ that constitute the \emph{cycle  graph} (Sec. \ref{sec_problem_description}), where $q^{xy}_1\in\mathbb{R}^2$ is associated to position $0$ in $1D$ and $L$ in $1D$ equals the sum of the lengths of edges connecting the $2D$  positions ($ \| q^{xy}_2 - q^{xy}_1 \| + \| q^{xy}_3 - q^{xy}_2 \| + \dots$). Given a $1D$ position $p_i(t)\in[0,L]$, its associated $2D$ position is obtained in two steps. First the particular edge associated to $p_i(t)$ is found, i.e., to find $s$ such that
\begin{align}
L_s <= p_i(t) <= L_{s+1},
\end{align}
where $L_s$ is the sum of the length of the edges from $q^{xy}_{1}$, up to the $2D$ $s-$th  position $q^{xy}_{s}$, 
\begin{align}
L_s&= \| q^{xy}_2 - q^{xy}_1 \| + \dots + \| q^{xy}_s - q^{xy}_{s-1} \|.
\end{align}
Second, the position $p^{xy}_i(t)$ on this edge between the $2D$ positions  
$q^{xy}_{s+1}$, $q^{xy}_{s}$, is found:
\begin{align}
p_i^{xy}(t) &= q^{xy}_{s} + (q^{xy}_{s+1}- q^{xy}_{s}) \frac{(p_i(t)- L_s)}{(L_{s+1}-L_s)}.
\end{align}
 Note that the Euclidean distance between any pair of $2D$ robot positions   $\|p^{xy}_{i+1}(t)- p^{xy}_{i}(t)\|$ on the cycle graph is always smaller than or equal to their distance on the $1D$ representation ($|p_i(t) - p_{i+1}(t)|$). Thus, if robots $i$ and $i+1$ are close enough in the $1D$ representation, this immediately means that they are also close enough in the Euclidean sense in what refers to their $2D$ positions. Thus, they can safely exchange data, as required by the method.

\begin{figure}[tbp]
\begin{center}
\begin{tabular}{c}
\includegraphics[width=0.32\paperwidth]{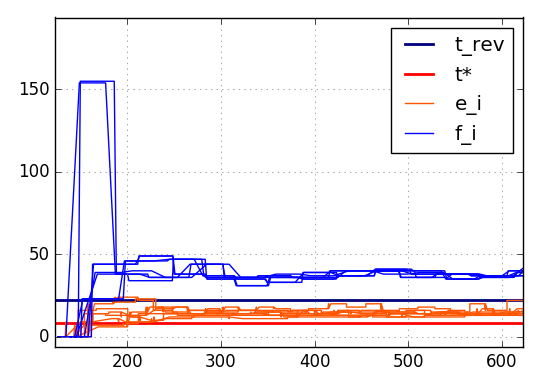}\\
{\footnotesize (a) $n=8$, $n_{bal}=4$}\\
\includegraphics[width=0.32\paperwidth]{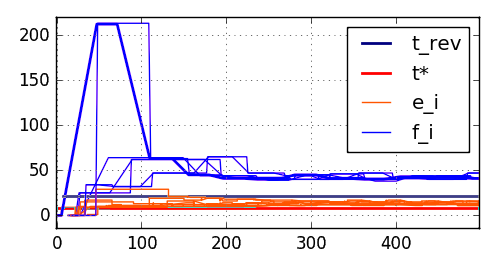}\\
{\footnotesize (b) $n=8$, $n_{bal}=3$}\\
\includegraphics[width=0.32\paperwidth]{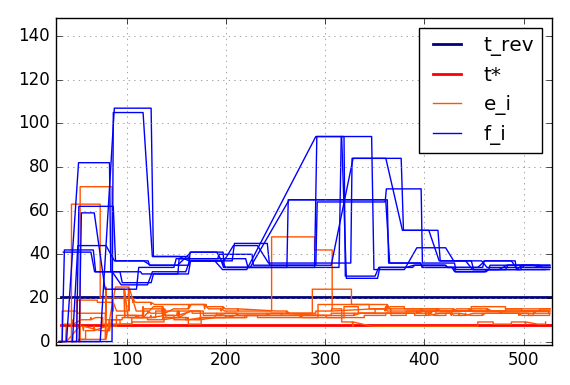}\\
{\footnotesize (c) $n=8$, $n_{bal}=4$. A robot decreases its speed (attached video).}
\end{tabular}
\end{center}
\caption{Evolution of $e_i(t)$ ~\eqref{eq_e_def} and $f_i(t)$ (Def.~\ref{def_revisiting_time}) compared  to $t_\star$ \eqref{eq_t_star_radii} and $t_{\mathrm{rev}}$ (Th.~\ref{th_performance_balanced} and \ref{th_performance_unbalanced}). 
(a): Balanced orientations with $n=8$. (b): Unbalanced case, with $n=8$ and $n_{bal}=3$. 
(c): Balanced orientations with $n=8$ robots, and 9 tasks. After some time, one robot decreases its speed and the robot team self--adapts to this new situation (attached video).
}
\label{fig_Gazebo_times}
\end{figure}

Fig. \ref{fig_Gazebo_environment} shows the original scenario. The task locations are numbered 1 to 8. The cycle graph used is the TSP associated to the task locations (in yellow, dashed). 
Robots travel the cycle graph until they reach a boundary, where they perform the needed calculations for the system to converge (Alg. \ref{alg_roboti_discovery}, \ref{alg_roboti_main} in Sec. \ref{sec_method}). This experiment is meant to check that no major problems arise when changing from a non-physical environment like Matlab into a more realistic one, and makes it one step closer to the possibility of real implementation. In the example in Fig. \ref{fig_Gazebo_environment}, the team is composed of 8 robots, with balanced  orientations. The results associated to this case are in Fig. \ref{fig_Gazebo_times}(a).

Robots run the proposed method (Sec. \ref{sec_method}, Algs. \ref{algo_discovery}, \ref{algo_main}). 
Figure \ref{fig_Gazebo_times} shows the results for a balanced configuration with $n=8$ robots (a), and for a team of $n=8$ robots with unbalanced orientations with $n_{bal}=3$ (b). Note that robots have physical restrictions (they cannot immediately stop or reach their maximum speeds, and their motions are no longer straight lines). Thus, as shown in Figure \ref{fig_Gazebo_times}, the traversing  $e_i(t)$  ~\eqref{eq_e_def} and revisiting times $f_i(t)$ (Def.~\ref{def_revisiting_time}) converge to values which are close to the theoretical $t_\star$ \eqref{eq_t_star_radii} and $t_{\mathrm{rev}}$  (Th.~\ref{th_performance_balanced} and \ref{th_performance_unbalanced}). These values are slightly  larger than the theoretical ones, due to the additional time employed to initiate and stop the motions, to turn around, and to exchange data. 

We also include an attached video with a simulation in Gazebo with $n=8$ robots and 9 tasks. Note that robots are not in charge of any fixed number of tasks. Instead, they are assigned to larger or smaller regions depending on their capabilities. In this example, after some time, one of the robots decreases its maximum speed. As a result,  traversing  $e_i(t)$  ~\eqref{eq_e_def} and revisiting times $f_i(t)$ (Def.~\ref{def_revisiting_time}) temporarily increase. The other team members self adapt to be in charge of larger regions, succeeding to decrease $e_i(t)$ and $f_i(t)$ after some iterations (Fig. \ref{fig_Gazebo_times} (c)). Thus, the proposed method 
succeeds to take advantage of the capabilities of all the robots in the team.

Additional videos with simulations in Gazebo (differential drive robots) and in  Matlab (ideal  robots) can be found at \url{https://www.youtube.com/playlist?list=PLmyvo-kjDwz30b_i8vW6gW0NeC3NHBvo6} and at \url{https://www.youtube.com/playlist?list=PL2pZRSxEnFj4AfrEbY1bjRUplmaSpZrBc}. These videos also include cases like the one in Fig. \ref{fig_Gazebo_environment}, but with $n=6$ robots instead of the $n=8$.

\section{Conclusions}
\label{sec_conclusions}

We presented a method to ensure a robotic network is kept intermittently connected. Robots move forward and backward on their regions, meeting intermittently with their previous and next neighbors. Simultaneously, they run a weighted consensus method to update their boundaries, so that the final regions associated to each robot can be traversed by them in a common time, that depends on the robots maximum speeds and communication radii. For balanced situations with the same number of robots moving forward and backward, robots converge to a configuration where the meetings take place simultaneously in the network and robots never wait at the region boundaries. For unbalanced situations, the performance degrades depending on how unbalanced the situation is. For this reason, future extensions of this  work include the design of distributed methods where robots change their orientations in order to get as close as possible to balanced configurations.

\appendices
\renewcommand{\thesectiondis}[2]{\Alph{section}:}

\section{Proof of Proposition \ref{prop_weightedConsensus}}
\label{appendices_prop_weightedConsensus}

In this section, we let matrix $V$, scalar  $\epsilon_i$, and 
matrices $P_i$, $\tilde{P}_i$, $\mathcal{L}_i$, $\tilde{\mathcal{L}}_i$ associated to the link $(i, i+1)$, be 
\begin{align}
V &= \mathrm{diag}(v_1, \dots, v_n), ~~~~   \epsilon_i = ~(v_i v_{i+1})/(v_i+v_{i+1}), \notag\\
P_i &= \mathbf{I} - (\mathrm{diag}(v_1, \dots, v_n))^{-1} \epsilon_i \mathcal{L}_i,
 \label{eq_P_i}\\
\tilde{P}_i &= V^{1/2}  P_i V^{-1/2} = \mathbf{I} - \tilde{\mathcal{L}}_i, 
~~~~\tilde{\mathcal{L}}_i= V^{-1/2}\epsilon_i \mathcal{L}_i V^{-1/2},\notag
\end{align}
where the entries in $P_i$, $\tilde{P}_i$, $\mathcal{L}_i$, and $\mathcal{\tilde{L}}_i$ are given by
\begin{align}
&[P_i]_{j,j'}= 0, ~\mathrm{except~for~} \label{eq_contents_P_i}\\
&[P_i]_{i,i}= 1- (\epsilon_i / v_i), 
~~~~[P_i]_{i+1,i+1}= 1- (\epsilon_i / v_{i+1}), \notag\\
&[P_i]_{i,i+1}= \epsilon_i / v_i, 
~~~~[P_i]_{i+1,i}= \epsilon_i / v_{i+1}, \notag\\
&[P_i]_{j,j}= 1, \mathrm{~for~all~} j\neq i, j\neq i+1,\notag
\end{align}
\begin{align}
&[\mathcal{L}_i]_{j,j'}= 0, ~\mathrm{except~for~} \label{eq_contents_L_i}\\
&[\mathcal{L}_i]_{i,i}= 1, [\mathcal{L}_i]_{i+1,i+1}= 1, 
[\mathcal{L}_i]_{i,i+1}= -1,
[\mathcal{L}_i]_{i+1,i}= -1, \notag
\end{align}
\begin{align}
&[\mathcal{\tilde{L}}_i]_{j,j'}= 0, ~\mathrm{except~for~} \label{eq_contents_tilde_L_i}\\
&[\mathcal{\tilde{L}}_i]_{i,i}= v_{i+1}/(v_i + v_{i+1}), 
~~[\mathcal{\tilde{L}}_i]_{i+1,i+1}= v_{i}/(v_i + v_{i+1}), \notag\\
&[\mathcal{\tilde{L}}_i]_{i,i+1}= [\mathcal{\tilde{L}}_i]_{i+1,i}=-\sqrt{v_i}\sqrt{v_{i+1}}/(v_i + v_{i+1}),\notag
\end{align}
\begin{align}
&[\tilde{P}_i]_{j,j'}= 0, ~\mathrm{except~for~} \label{eq_contents_tilde_P_i}\\
&[\tilde{P}_i]_{i,i}= 1- \frac{v_{i+1}}{v_i+v_{i+1}},
~~[\tilde{P}_i]_{i+1,i+1}= 1- \frac{v_{i}}{v_i+v_{i+1}}, \notag\\
&[\tilde{P}_i]_{i,i+1}=[\tilde{P}_i]_{i+1,i}= \sqrt{v_i}\sqrt{v_{i+1}}/(v_i + v_{i+1}), \notag\\
&[\tilde{P}_i]_{j,j}= 1, \mathrm{~for~all~} j\neq i, j\neq i+1.\notag
\end{align}

\begin{lemma}\cite{Aragues-19ACC}
\label{lemma_eigs_Pi}

The eigenvalues of the matrices $P_i$, $\tilde{P}_i$ defined in ~\eqref{eq_P_i},  for all $(i,i+1)$ links, with $i=1,\dots,n-1$, satisfy:
\begin{align}
&\lambda(P_i) \in (-1,1] &&\lambda(\tilde{P}_i) \in (-1,1].
\label{eq_lemma_eigs_Pi}
\end{align}
\end{lemma}
\begin{IEEEproof}

The eigenvalues of $\tilde{P}_i$ are
\begin{align}
\lambda(\tilde{P}_i)&= 1- \lambda(V^{-1/2}\epsilon_i \mathcal{L}_i V^{-1/2}) = 1-\lambda(\tilde{\mathcal{L}}_i).
\label{eq_eigs_tildeP_tildeL}
\end{align}

Note that $\mathcal{L}_i$ \eqref{eq_P_i}, \eqref{eq_contents_L_i} is the unweighted symmetric Laplacian matrix associated to the link $(i, i+1)$, and thus it is positive semidefinite \cite{olfati2007consensus}. Since $\epsilon_i > 0$, and matrix $V^{-1/2}$ is positive definite and symmetric, then $\tilde{\mathcal{L}}_i=\lambda(V^{-1/2}\epsilon_i \mathcal{L}_i V^{-1/2})$ \eqref{eq_P_i} \eqref{eq_contents_tilde_L_i}  is positive semidefinite ~\cite[Chapter~7.1]{horn1990matrix}, with eigenvalues larger than or equal to $0$, and with its largest eigenvalue begin smaller than or equal to the infinite matrix norm $\|\tilde{\mathcal{L}}_i\|_\infty=\max_j (|[\tilde{\mathcal{L}}_i]_{j1}|+\dots+|[\tilde{\mathcal{L}}_i]_{jn}|)$,
\begin{align}
&\lambda_{\max}(\tilde{\mathcal{L}}_i)\leq \|\tilde{\mathcal{L}}_i\|_\infty 
= \frac{\max \left(  v_{i} , v_{i+1} \right) + \sqrt{v_i}\sqrt{v_{i+1}}}{v_i+v_{i+1}}
\label{eq_eits_tildeL}\\
&\leq (\max \left(  v_{i} , v_{i+1} \right) + \max\left(v_i, v_{i+1}\right))/(v_i+v_{i+1})< 2.\notag
\end{align}
From eqs.~\eqref{eq_eigs_tildeP_tildeL},\eqref{eq_eits_tildeL}, for all $i=1,\dots,n-1$,
\begin{align}
-1 = (1-2) < \lambda(\tilde{P}_i) \leq (1-0)=1,
\end{align}
and since $P_i$ is \emph{similar} to $\tilde{P}_i$  \eqref{eq_P_i}, then $P_i$ and $\tilde{P}_i$ have the same eigenvalues, and  thus we conclude \eqref{eq_lemma_eigs_Pi}.
\end{IEEEproof}

\begin{proposition}\cite{Aragues-19ACC}
\label{prop_weighted_consensus}

Let matrices $P_{i(t)}$, $\tilde{P}_{i(t)}$ be as in \eqref{eq_P_i}. If the set of matrices that appear infinitely often are jointly connected, then, for all $z(0)$, $e(0)$, the iterations 
 $z(t^+) = \tilde{P}_{i(t)} z(t)$,  $e(t^+) = P_{i(t)} e(t)$, with 
\begin{align}
z(t)&=V^{1/2} e(t), &e(t)&=V^{-1/2} z(t), 
\label{eq_relation_z_e}
\end{align}
and $V$ as in \eqref{eq_P_i}, 
converge respectively to 
\begin{align}
z_\star &= (V^{1/2} \mathbf{1} \mathbf{1}^T V^{1/2} / \mathbf{1}^T V \mathbf{1} ) z(0), \label{eq_z_star}\\
e_\star &= (\mathbf{1} \mathbf{1}^T / \mathbf{1}^T V \mathbf{1}) V e(0).\notag
\end{align}

\end{proposition}
\begin{IEEEproof}
Consider the matrix $\tilde{P}_{i_1:i_{j}}$ associated to a particular jointly connected sequence  $i_j  \dots i_1$ (the sequence takes place in the opposite order to matrix multiplication),
\begin{align}
\tilde{P}_{i_1:i_{j}} &= \tilde{P}_{i_1} \tilde{P}_{i_2} \dots \tilde{P}_{i_j}. 
\end{align}
For this matrix, 
\begin{align}
\rho(\tilde{P}_{i_1:i_{j}} ) &\leq \|\tilde{P}_{i_1:i_{j}} \|_2 \leq 
\| \tilde{P}_{i_1} \|_2 \|\tilde{P}_{i_2} \|_2 \dots \|\tilde{P}_{i_j}  \|_2 \notag\\
&= \rho(\tilde{P}_{i_1}) \rho(\tilde{P}_{i_2}) \dots \rho(\tilde{P}_{i_j}) = 1,
\label{eq_rhos_jointly}
\end{align}
where we have used Lemma \ref{lemma_eigs_Pi} ($\lambda(\tilde{P}_i) \in (-1,1]$ for all $i=1,\dots,n-1$), and the fact that $\tilde{P}_{i_1}, \tilde{P}_{i_2}, \dots,  \tilde{P}_{i_j}$ are symmetric, and their spectral norms equal their spectral radius,
$
\|\tilde{P}_i\|_2 = \rho(\tilde{P}_i) = \max(|\lambda(\tilde{P}_i)  |)$. Thus, all the eigenvalues of matrix $\tilde{P}_{i_1:i_{j}}$ are between $[-1, +1]$. 

Now we pay attention to the structure of matrix $\tilde{P}_{i_1:i_{j}}$. Every matrix $\tilde{P}_i$ \eqref{eq_contents_tilde_P_i} has all the entries equal to zero, apart from the diagonal terms $(1,1)\dots(n,n)$, and the entries $(i,i+1)$, $(i+1,i)$, which are strictly positive. 
After multiplying matrices $\tilde{P}_{i_1}, \dots, \tilde{P}_{i_j}$, we get a nonnegative matrix $\tilde{P}_{i_1:i_{j}}$ that has \emph{at least} the following elements strictly positive (the remaining entries may be zero or positive): the diagonal terms $(1,1)\dots(n,n)$, and all the $(i, i+1)$ and $(i+1, i)$ entries associated to the $i, i+1$ links that appear in each associated matrix $P_{i}$, for $i=i_1\dots i_j$. Matrix $\tilde{P}_{i_1:i_{j}}$ contains at least all matrices associated to the $n-1$ different links   ($i,i+1$), for all $i=1,\dots,n-1$. 
Then, the structure of $\tilde{P}_{i_1:i_{j}}$ contains \emph{at least} positive  elements in all the entries $(i,i)$ for $i=1,\dots,n$, and $(i,i+1)$, $(i+1,i)$ for $i=1,\dots, n-1$. Thus, matrix $\tilde{P}_{i_1:i_{j}}$ is \emph{primitive} and \cite{jadbabaie2003coordination}, \cite{horn1990matrix} among its $n$ eigenvalues, there is exactly one with the largest magnitude, and this eigenvalue is the only one possessing  an eigenvector with all positive entries, and the remaining $n-1$ eigenvalues are all strictly smaller in magnitude than the largest one. 
From \eqref{eq_rhos_jointly}, this eigenvalue has modulus smaller than or equal to 1. 

Now note that for each matrix $P_i$ \eqref{eq_P_i},
\begin{align}
P_i \mathbf{1}&=  \mathbf{1}, ~~~~~~~~\mathrm{and~that}\notag\\
\tilde{P}_{i_1:i_{j}} &= \tilde{P}_{i_1} \tilde{P}_{i_2} \dots \tilde{P}_{i_j}
= V^{1/2} P_{i_1} P_{i_2} \dots P_{i_j} V^{-1/2}.
\label{eq_eigenvects_tmp}
\end{align}
From \eqref{eq_eigenvects_tmp}, we conclude that $V^{1/2}\mathbf{1}$ is the eigenvector of  $\tilde{P}_{i_1:i_{j}}$  associated to the eigenvalue $1$,
\begin{align}
\tilde{P}_{i_1:i_{j}} V^{1/2}\mathbf{1} = V^{1/2}\mathbf{1}.
\end{align}
This eigenvector has all its entries positive, and it is associated to the largest modulus eigenvalue, which has to be $1$ and not $-1$. 

Matrix $\tilde{P}_{i_1:i_{j}}$ is also paracontractive (e.g., \cite[Corollary~2]{calafiore2009distributed}, using $\mathrm{span}(V^{1/2}\mathbf{1})$ instead of $\mathrm{span}(\mathbf{1})$).

From \cite[Theorem~1]{elsner1990convergence} \cite[Theorem~2]{xiao2005scheme}: suppose that a finite set of square matrices $\{W_1, \dots, W_j\}$ are paracontractive, and denote $\mathcal{J}$ the  set of integers that appear infinitely often in the sequence. Then, for all $\tilde{z}(0)$, the sequence of vectors $\tilde{z}(t_e^+) = W_{i(t_e)} \tilde{z}(t_e)$ has a limit $z_\star \in \cap_{i\in\mathcal{J}} \mathcal{H}(W_i)$, with $\mathcal{H}(W_i)=\{z | W_i z = z\}$. In our case, we use the fact that all our possible jointly connected matrices have the  common eigenvector $V^{1/2}\mathbf{1}$ associated to the eigenvalue 1 and it is the only one. Thus, $\tilde{z}(t_e^+) = \tilde{P}_{i_1:i_j(t_e)}  \tilde{z}(t_e)$, and thus $z(t_e^+) = \tilde{P}_{i(t_e)} z(t_e)$, converge to $z_\star$ in \eqref{eq_z_star}. Due to \eqref{eq_relation_z_e}, $e(t_e^+)=P_{i(t_e)} e(t_e)$ converges to $e_\star$ \eqref{eq_z_star}.
\end{IEEEproof}

Using the previous intermediary results, we are ready to prove Proposition \ref{prop_weightedConsensus}.

\begin{IEEEproof}[Proof of Proposition \ref{prop_weightedConsensus}]
We first consider how the region lengths $d_i(t)$, $d_{i+1}(t)$ (eq.~\eqref{eq_d_def})  evolve when robots $i$, $i+1$, $i\in{1,\dots,n-1}$, update their boundary $y_i(t_e^+)$ with eq.~\eqref{eq_interMeetingPoint_update_radii} (the other region lengths are not affected by \eqref{eq_interMeetingPoint_update_radii}). If several simultaneous events take place, we will consider any ordering, e.g., first the ones with lower identifiers. Note that \eqref{eq_region_length_d_update_1} will be the same in the presence of simultaneous updates, since it depends on boundaries which require actions from robots $i$ and $i+1$ and,  since they are currently involved in their meeting at the common boundary $y_i(t)$, they cannot be simultaneously involved in other meetings at the other boundaries (i.e., robot $i$ is not at boundary $y_{i-1}(t)$, and robot $i+1$ is not at $y_{i+1}(t)$),
\begin{align}
d_{i}(t_e^+)&=y_{i}(t_e^+)-y_{i-1}(t_e^+)=y_{i}(t_e^+)-y_{i-1}(t_e), \label{eq_region_length_d_update_1}\\
d_{i+1}(t_e^+)&=y_{i+1}(t_e^+)-y_{i}(t_e^+)=y_{i+1}(t_e)-y_{i}(t_e^+). \notag
\end{align}
After some manipulation, \eqref{eq_region_length_d_update_1} is equivalent to:
{\small
\begin{align}
&d_{i}(t_e^+)=\frac{v_{i} (d_{i+1}(t_e) + d_{i}(t_e))}{v_i+v_{i+1}} + 2\frac{v_{i+1}r_i - v_i r_{i+1}}{v_i + v_{i+1}}, \label{eq_region_length_d_update_2}\\
&d_{i+1}(t_e^+)=\frac{v_{i+1} (d_{i+1}(t_e) + d_{i}(t_e))}{v_i+v_{i+1}} -2\frac{v_{i+1}r_i- v_i r_{i+1}}{v_i + v_{i+1}}. \notag
\end{align}
}
The traversing times $e_{i}(t)$, $e_{i+1}(t)$ (eq.~\eqref{eq_e_def}) of robots $i$ and $i+1$ are also affected by ~\eqref{eq_interMeetingPoint_update_radii}, due to \eqref{eq_region_length_d_update_2} (the other traversing times are not affected by \eqref{eq_interMeetingPoint_update_radii}):
\begin{align}
&e_{i}(t_e^+)=e_{i}(t_e) + \frac{\epsilon_i}{v_{i}} (e_{i+1}(t_e) -e_{i}(t_e) ), \mathrm{~} \epsilon_i = \frac{v_i v_{i+1}}{v_i+v_{i+1}}, \notag\\
&e_{i+1}(t_e^+)=e_{i+1}(t_e) - \frac{\epsilon_i}{v_{i+1}} (e_{i+1}(t_e) - e_{i}(t_e)). \label{eq_region_time_e_update_2_app}
\end{align}

In matrix form, eq.~\eqref{eq_region_time_e_update_2_app} is a discrete--time switching  weighted consensus, with Perron matrix \cite{olfati2007consensus} $P_i$ as in  \eqref{eq_P_i} found in Appendix \ref{appendices_prop_weightedConsensus},
\begin{align}
e(t_e^+)&=P_{i(t_e)} e(t_e), &e(t)&=\left[e_{1}(t), \dots, e_{n-1}(t)\right]^T.
\label{eq_update_e_vect}
\end{align}
From Proposition \ref{prop_weighted_consensus} in Appendix \ref{appendices_prop_weightedConsensus}, 
 if the set of communication graphs that occur infinitely often are jointly connected, then $e(t)$ in  \eqref{eq_update_e_vect} converges to $e_{\star}$ in \eqref{eq_z_star}. Thus, for all $i=1,\dots,n$, $e_{i}(t)$ defined by eq.~\eqref{eq_e_def} and evolving as in  \eqref{eq_region_time_e_update_2_app}, converge to the weighted average of $e_{j}(0)$, with weighting vector given by $[v_1, \dots, v_n]^T$ (Proposition \ref{prop_weighted_consensus} in the Appendix and eq. \eqref{eq_P_i}),
{\small
\begin{align}
\mathop{\lim}_{t \rightarrow \infty} e_{i}(t) 
= \frac{\mathop{\sum}_{j=1}^n v_j e_{j}(0) }{v_1+\dots+v_n}
= \frac{\mathop{\sum}_{j=1}^n \frac{v_j(d_j(0)-2r_j)}{v_j} }{v_1+\dots+v_n}
 = t_{\star},
\end{align}
}
since $d_1(0)+\dots+d_n(0)=y_n(t)=L$ (eq.~\eqref{eq_d_def}), with $t_\star$ as in ~\eqref{eq_t_star_radii}, 
and thus, the region lengths $d_i(t)$,  and the boundaries $y_i(t)$ converge to the values in ~\eqref{eq_d_star_radii}, \eqref{eq_optimal_boundary}. 
\end{IEEEproof}

\section{Proofs of Lemmas \ref{lemma_STmove}, \ref{lemma_discrete_asynchronous_behavior} and \ref{lemma_properties}}
\label{appendices_lemmas_weigthedConsensus}

\begin{IEEEproof}[Proof of Lemma \ref{lemma_STmove} \cite{Aragues-19ACC}]
Since $L$ is fixed, and $y_n(t)=L$ is fixed then, for all $i=1,\dots,n$, $d_i(t)\leq L$ and thus $e_i(t)\leq L/v_i$ \eqref{eq_e_def}. Active robots move from a position inside their  region to one of their boundaries, employing thus a time less or equal to $e_i(t)$, which as we saw, is bounded. 
After that, the event is an \emph{arrival} if the boundary is empty, and a \emph{meeting} if the neighbor is already waiting at the boundary.
\end{IEEEproof}

\begin{IEEEproof}[Proof of Lemma \ref{lemma_discrete_asynchronous_behavior} \cite{Aragues-19ACC}]
The proof follows by inspection of Def.~\ref{def_discreteAsynchVersion} and Algorithm~\ref{algo_main}.
\end{IEEEproof}

\begin{IEEEproof}[Proof of Lemma \ref{lemma_properties} \cite{Aragues-19ACC}] 

$(i)$ \emph{About the orientations}: Orientations $o_i(t)$ only change during \emph{discovery} and \emph{meeting} events and, in both cases (eqs.~\eqref{eq_after_discovery}, \eqref{eq_after_meeting}), the orientations of the two  involved agents $i$ and $i+1$ are simultaneously changed, and thus the sum of all orientations remains unchanged.
 
$(ii)$ \emph{About the regions being disjoint, with the only common point being the boundary}: This is true during the \emph{discovery} and \emph{catch}, where robots $i$, $i+1$ define their common boundary at the same time. After, at meetings (eq.~\eqref{eq_interMeetingPoint_update_radii}) robots $i$, $i+1$ change simultaneously their common boundary $y_i(t_e^+)$, and the update rule makes $y_i(t_e^+)$ remain  strictly between $y_{i-1}(t_e^+)=y_{i-1}(t_e)$ and $y_{i+1}(t_e^+)=y_{i+1}(t_e)$.

$(iii)$ \emph{About robots not exchanging positions}: The region associated to each  robot $i$ is defined by the boundaries $y_{i-1}(t)$ and $y_{i}(t)$. After meeting with  robot $i+1$, the position of $y_i(t_e^+)$ changes (eq.~\eqref{eq_interMeetingPoint_update_radii}), with $y_i(t_e^+)\in [y_{i-1}(t_e), y_{i+1}(t_e)]$. Then, robot $i$ goes to its other boundary  $y_{i-1}(t)\leq y_i(t)$, and when later it comes back to boundary $y_i(t)$, it holds $y_i(t)\leq y_{i+1}(t)$, regardless of the fact that robot $i+1$ may have updated or not $y_{i+1}(t)$. Thus, robot $i$ will reach $y_i(t)$ and will never reach $y_{i+1}(t)$. Thus, in the discrete asynchronous behavior (Def.~\ref{def_discreteAsynchVersion}), robots do not exchange the positions.
\end{IEEEproof}

\section{Proof of Proposition \ref{prop_jointconn}}
\label{appendices_prop_jointconn}

\begin{IEEEproof}[Proof of Proposition \ref{prop_jointconn}]
Considering the discrete asynchronous behavior (Def.~\ref{def_discreteAsynchVersion}) of the algorithm, we represent the system as $n$ boundaries and $n$ robots placed at the boundaries. As we show next, the system cannot experience blocking, and the  meetings propagate through the network.

\noindent \emph{Blocking}: The only possible blocking situation is one with each robot waiting  at a different boundary since, as long as two robots fall in a common  boundary, a \emph{meeting} event takes place, making the system evolve. At the blocking, robots with $o_i(t)>0$ would be at their right boundary, and robots with  $o_i(t) < 0$ at their left boundary, with these boundaries being the unique common points between the disjoint regions of each robot (Lemma~\ref{lemma_properties}$(ii)$). From Assumption $(A2)$, at least one robot has a  different initial orientation than the others, and by Lemma~\ref{lemma_properties} $(i)$ this continues during the whole execution of the method. Thus, at least two robots will reach the same boundary, giving rise to a \emph{meeting} event.

\noindent \emph{Propagation}: After robots $i$, $i+1$ meet, they move to their opposite  boundaries, thus propagating the process to the preceding and following robots  $i-1$ and $i+2$, since the order of the robots is preserved (Lemma~\ref{lemma_properties} $(iii)$). Repeating the same reasoning with robots $i-1$, $i+2$, we conclude that meetings are propagated through the network. The only reasons not to propagate would be  either a blocking (we proved this is not possible), or that the same subset of robots would be meeting each other, without involving the others. But in order for $i,i+1$ to meet again, there must have been a meeting between  $i-1,i$ and $i+1, i+2$, so this case is discarded as well.

\noindent \emph{Bounded times}: The discrete asynchronous behavior (Def.~\ref{def_discreteAsynchVersion}) does not exhibit blocking and ensures propagation of the meetings, and the time between events is bounded (Lemma \ref{lemma_STmove}). Therefore, the time interval required for the network to be jointly connected is bounded.
\end{IEEEproof}

\section{Proofs of Lemmas \ref{lemma_rounds} and \ref{lemma_interlaced}}
\label{appendices_lemmas_Rounds_interlaced}

\begin{IEEEproof}[Proof of Lemma \ref{lemma_rounds}]
\noindent $(i)$ Let $t_{e}$ be the time at which the meeting between robots $i$,  $i+1$ takes place within the current round $k$, with 
\begin{align}
t_0 + k t_\star \leq t_{e} < t_0 + (k+1) t_\star.
\label{eq_proof_lemma_rounds_i}
\end{align}
Note that the time $t_e$ of the meeting of robots $i$, $i+1$ is associated to  the last robot that arrived to the boundary, so that $t_e=\max\{t^e_i(k), t^e_{i+1}(k)\}$ in \eqref{eq_dis_synch_upd}. From \eqref{eq_discreteAsynchVersion_event_time}, with $e_i(t_e^+)=t_\star$ (Asm. \ref{ass_commonTraversingTimes}), the arrivals of robots $i$, $i+1$ to the opposite boundaries will take place at time 
\begin{align}
t_{e'}=t_{e}+ t_\star=\max\{t^e_i(k), t^e_{i+1}(k)\}+t_\star,
\end{align}
giving $t_{e'}=t^e_i(k+1)=t^e_{i+1}(k+1)$ in \eqref{eq_dis_synch_upd}. From eq.~\eqref{eq_proof_lemma_rounds_i}, these arrival times $t_{e'}=t^e_i(k+1)=t^e_{i+1}(k+1)$ take place during the round $k+1$:
\begin{align}
t_0 + (k+1) t_\star \leq t_{e}+ t_\star < t_0 + (k+2) t_\star.
\end{align}
Since between the meeting time $t_e$ at round $k$ and the arrival time $t_{e'}$ at round $k+1$ robots $i$ and $i+1$ are traveling between boundaries, they cannot be involved in any  additional event (arrivals or meetings). This concludes the proof of $(i)$.

\noindent $(ii)$ Due to the previous property, a robot $i$ cannot be involved in  more than a meeting during the current round. Since meetings affect at most the two robots involved, their state updates can be performed independently of the state updates of the other robots involved in meetings during the current round. The proof is completed by considering this observation, together with the equivalence of  the event times discussed in the proof of $(i)$, and the definition of the Discrete Synchronous Behavior (Def.~\ref{def_behaviorDiscreteSynch} and eq.~\eqref{eq_dis_synch_upd}).

\noindent $(iii)$ Both states are simultaneously updated \eqref{eq_dis_synch_upd}, and both represent the fact that, during round  $k$, both robots will be at or arrive to the common boundary $y^\star_i$ and a meeting will take place. 
\end{IEEEproof}

\begin{IEEEproof}[Proof of Lemma \ref{lemma_interlaced}]

At the beginning of round $k$, orientations are interlaced as in Definition~\ref{def_interlaced} with $n_+ \geq n_-$. 

Then, during the current round $k$, all robots with indexes $i_j$, for all $j=1,\dots,n_{bal}$, as in \eqref{eq_interlaced_index}, will arrive to the boundary $y_{i_j}^\star$, and all robots with indexes $i_j+1$ will arrive to the boundary $y_{i_j}^\star$. Thus, during the current round $k$ there will be $n_{bal}$ meetings at these boundaries, involving $2 n_{bal}$ robots. Due to these meetings (eq. \eqref{eq_dis_synch_upd}), the involved robots 
will arrive to the opposite boundaries at round $k+1$ and will change their orientations (Lemma~\ref{lemma_rounds}),
i.e., for all $j=1,\dots,n_{bal}$: 
\begin{align}
o_{i_j}(k+1)=-1 ~\mathrm{and}~ o_{i_j + 1}(k+1)=+1.
\label{eq_tmp_proof_interlacemente_conservance}
\end{align}
Since $n_+ \geq n_-$ and, from Lemma~\ref{lemma_properties}, $n_+$, $n_-$, $n_{bal}$ remain unchanged for all rounds and  times, then the orientations of the remaining robots at round $k+1$ are positive. In particular, $o_{i_1 -1}=+1$, and from \eqref{eq_tmp_proof_interlacemente_conservance}, during round $k+1$,   
the orientations are interlaced with robot indexes $i_1-1, i_2-1, \dots, i_{n_{bal}}-1$, i.e., 
for all $j=1,\dots,n_{bal}$:
\begin{align}
o_{i_j -1 }(k+1)=+1 ~\mathrm{and}~ o_{i_j}(k+1)=-1.
\end{align}
The proof is completed by repeating the reasoning for the successive rounds. 
\end{IEEEproof}

\section{Proofs of Propositions \ref{prop_performance_unbalanced_interlaced} and \ref{prop_balanced_interlaced_clock_synch}}
\label{appendices_props_interlaced_performance_synch}

\begin{IEEEproof}[Proof of Prop. \ref{prop_performance_unbalanced_interlaced}]
It is a consequence of Lemma~\ref{lemma_interlaced}: each robot $i$ arrives to the $y_i^\star$ boundary $n_{bal}$ times every $n$ rounds. And from Assumption ~\ref{ass_commonTraversingTimes} and Def. ~\ref{def_round}, the duration of each round equals $t_\star$.
\end{IEEEproof}

\begin{IEEEproof}[Proof of Prop. \ref{prop_balanced_interlaced_clock_synch}]
We use Lemma~\ref{lemma_interlaced} with $n_{bal}=n/2$ since orientations are balanced. Since at round $k_0$ the orientations are balanced interlaced, from Lemma~\ref{lemma_interlaced} they remain balanced  interlaced for all the successive rounds. At each round $k$ there are exactly $n/2$ meetings involving $n$ robots. Thus, if at round $k$ the robot indexes are ${i_1,i_2,\dots,i_{n/2}}=\{1,3,\dots,n-1\}$, at round $k+1$ the robot indexes are ${i_1-1,i_2-1,\dots,i_{n/2}-1}=\{0,2,\dots,n-2\}$, i.e., due to the cycle structure they equal to $\{2,\dots,n-2,n\}$. Thus, every robot $i$ meets at every round with one of its neighbors, and with the opposite neighbor during the next round.

Now consider the arrival of robots $i$ and $i+1$  at the common boundary $y_i^\star$ at round $k$, which takes place at times $t_i^{e}(k)$ and $t_{i+1}^{e}(k)$. This will give rise to a meeting during the current round $k$ that will take place at time $\max\{t_i^{e}(k), t_{i+1}^{e}(k)  \}$. Equivalently, this will give rise to two arrivals to boundaries (the opposite ones) that will take place at times \eqref{eq_dis_synch_upd}
\begin{align}
t_i^{e}(k+1)=t_{i+1}^{e}(k+1)&=\max\{t_i^{e}(k), t_{i+1}^{e}(k)  \} + t_\star.
\label{eq_balanced_interlaced_clock_upd}
\end{align}
Now note that, at every round, every robot $i$ meets exactly once, so that eq.~\eqref{eq_balanced_interlaced_clock_upd} adds $t_\star$ to every $t_i^{e}(k)$ at every round. 
Note also that during round $k$, robot $i$ met with robot $i-1$, and robot $i+1$ met with robot $i+2$. Thus, \eqref{eq_balanced_interlaced_clock_upd} gives
\begin{align}
t_i^{e}(k+1)&=2 t_\star + \mathop{\max}_{j=i-1}^{i+2} \{t_j^{e}(k-1)\}.
\label{eq_balanced_interlaced_clock_upd_2}
\end{align}
During round $k-1$, the states of the robots involved in \eqref{eq_balanced_interlaced_clock_upd_2} were updated due to meetings between robots $(i-2,i-1)$, $(i,i+1)$, $(i+2,i+3)$, so that
\begin{align*}
t_i^{e}(k+1)&=3 t_\star + \mathop{\max}_{j=i-2}^{i+3} \{t_j^{e}(k-2)\},
\end{align*}
and, for $k$ large enough, the reasoning can be repeated until it is propagated to the set of initial values $t_j^{e}(k_0)$. 
 Thus, we can see that \eqref{eq_balanced_interlaced_clock_upd} is equivalent to a $\max$ consensus \cite{Tahbaz-Salehi-CDC2006} on the initial values $t_1^{e}(k_0), \dots, t_n^{e}(k_0)$, plus the term $(k-k_0) t_\star$ which places the event within the current round. Due to the graph structure of the network, its diameter equals $n/2$, and thus, after $n/2$ iterations the $\max$ consensus converges to the maximal initial value, and equivalently, all the time events associated to the arrivals take place at:
\begin{align}
t_i^{e}(k)&=(k-k_0) t_\star + \max_j\{ t_j^{e}(k_0) \}, ~\mathrm{for~all}~i=1,\dots,n.
\end{align}
Thus, from each pair of simultaneous arrivals, the meeting takes place immediately. From this observations, we get the revisiting times $2 t_{rev} = 2 t_\star$.
\end{IEEEproof}

\section{Proofs of Lemmas \ref{lemma_sequence_evolution_rules}, \ref{lemma_sequence_properties} and \ref{lemma_collaborative_sequences_reduce}}
\label{appendices_lemmas_sequences_basic}

\begin{IEEEproof}[Proof of Lemma \ref{lemma_sequence_evolution_rules}]
From  ~\eqref{eq_dis_synch_upd}, at every round $k+1$, all the letters in $w(k)$ remain the same in $w(k+1)$, and all the sequences in $w(k)$ reverse their values ($+$ become $-$ and $-$ become $+$) due to the meetings that take place at round $k$. 
The previous rules follow from direct application of this fact. 
\end{IEEEproof}

\begin{IEEEproof}[Proof of Lemma \ref{lemma_sequence_properties}]
($i$): The proof is immediate, by looking at the rules. 
($ii$): The creation of a new sequence requires $+-$ to appear at round $k+1$. By inspection of the rules, this cannot happen, neither by the individual application of the rules, or by the interactions between the rules of different sequences.
\end{IEEEproof}

\begin{IEEEproof}[Proof of Lemma \ref{lemma_collaborative_sequences_reduce}]
From Lemmas~\ref{lemma_sequence_evolution_rules},~\ref{lemma_sequence_properties}, sequences move at most one position per round and, a sequence that experiences a \emph{Reduce} rule, generates letters $-$ and $+$ around it which cannot be consumed in a single round by other sequences. Thus, in round $k+1$ it will \emph{Reduce} again, and again at $k+2, \dots$ until it disappears. 
\end{IEEEproof}

\section{Proof of Lemma \ref{lemma_collaborative_sequences_move_neg}}
\label{appendices_lemma_move_neg}

\begin{IEEEproof}[Proof of Lemma \ref{lemma_collaborative_sequences_move_neg}]
Consider sequence $[seq,l]$ experiences the \emph{Move-} rule and there are additional $-$ letters to its right. This sequence  always introduces a $-$ letter to the left. Since every other sequence $[seq',l']$ to the left of $[seq,l]$ cannot move faster   than 1 position per round (Lemma ~\ref{lemma_sequence_properties}) then, even if this other sequence consumes this recent letter, it cannot consume it faster than sequence $[seq,l]$ creates it. Thus, as long as a sequence $[seq_j,l]$ experiences the \emph{Move-} rule for the  first time, it keeps on running the \emph{Move-} rule while there are $-$ letters to the right.  

We consider sequence $[seq,l]$ has already consumed all the extra $-$ letters to its right, placing them to its left. The different situations that it can find are described and analyzed next:

\begin{align*}
(a)~ \mathrm{if~}  w(k)_{j1: j1+l+l'+2}=~ & - [seq, l] - [seq',l']- \\
 \mathrm{then ~}  w(k+1)_{j1: j1+l+l'+2}=~ &   (-) - [seq, l] - [seq',l']
\end{align*}
Both sequences \emph{Move-} to the right, until something happens to $[seq',l']$ that makes $[seq,l]$ act accordingly. Since $n_+ \geq n_-$, at some point the sequence $[seq',l']$ will meet a $+$  and will act as described in case $(b)$.  

\begin{align*}
(b)~ \mathrm{if~}  w(k)_{j1: j1+l+l'+2}=~ &- [seq, l] - [seq',l']+   ~ \mathrm{then}\\
  w(k+1)_{j1: j1+l+l'+2}=~ &   (-) -[seq, l] - [seq',l'-2]+ (+)
\end{align*}
Sequence $[seq',l']$ \emph{Reduces}, feeding with additional $-$ to $[seq,l]$ until $[seq',l']$  disappears (Lemma~\ref{lemma_collaborative_sequences_reduce}) and $[seq,l]$ finds the first $+$. This case $(e)$ is explained later. 

\begin{align*}
(c)~ \mathrm{if~}  w(k)_{j1: j1+l+l'+3}=&&& - [seq, l]- + [seq',l']- \\
\mathrm{then~}w(k+1)_{j1: j1+l+l'+3}=&&&  (-) - [~seq',  l' + l + 2~]
\end{align*}
Sequence $[seq',l']$ \emph{Expands}, $[seq,l]$ \emph{Move-}, and then both \emph{Merge} in a larger sequence. Note this merged sequence has a $-$ to the left. Thus (Lemma~\ref{lemma_sequence_evolution_rules}), it will  either \emph{Reduce} (discussed in Lemma~\ref{lemma_collaborative_sequences_reduce}) or start performing \emph{Move-} (behaving according to the different cases discussed in this proof). 

\begin{align*}
(d)~ \mathrm{if~}  w(k)_{j1: j1+l+l'+3}= &&& - [seq, l]- + [seq',l']+ \\
\mathrm{then~}w(k+1)_{j1: j1+l+l'+3}=&&&  (-) - [seq',  l' + l] + (+)
\end{align*}
Sequence $[seq,l]$ \emph{Move-}, $[seq',l']$ \emph{Move+}, and then both \emph{Merge} in a larger sequence. Since this sequence has a letter $-$ to the left and $+$ to the right, then it will \emph{Reduce}, behaving as discussed in Lemma~\ref{lemma_collaborative_sequences_reduce}. 

\begin{align*}
(e)~ \mathrm{if~}  w(k)_{j1: j1+l+3}= &&& -~[seq, l]~-+~+\\
\mathrm{then~} w(k+1)_{j1: j1+l+3}= &&& (-)-[seq, l]+(+)\\ 
\end{align*}
Finally, we consider that sequence $[seq,l]$ has some $+$ letters to the right that do not belong to sequences (thus, there are at least two consecutive $+$ letters). It makes a \emph{Move-}, and finishes with a $-$ to the left and $+$ to the right. Then, it will \emph{Reduce}, behaving as discussed in Lemma~\ref{lemma_collaborative_sequences_reduce}.
\end{IEEEproof}

\section{Proof of Lemma \ref{lemma_collaborative_sequences_move_pos}}
\label{appendices_lemma_move_pos}

\begin{IEEEproof}[Proof of Lemma \ref{lemma_collaborative_sequences_move_pos}]
The proof is organized by considering each case separately.

Consider sequence $[seq,l]$ experiences the \emph{Move+} rule and there are additional $+$ letters to its left. This sequence  always introduces a $+$ letter to the right. Since every other sequence $[seq',l']$ to the right of $[seq,l]$ cannot move faster  than 1 position per round (Lemma ~\ref{lemma_sequence_properties}) then, even if this other sequence consumes this recent letter, it cannot consume it faster than sequence $[seq,l]$ creates it. Thus, as long as a sequence $[seq_j,l]$ experiences the \emph{Move+} rule for the  first time, it keeps on running the \emph{Move+} rule while there are $+$ letters to its left, placing them to its right.

When the orientations are unbalanced, with $n_+ > n_-$, it may be the case that there are always $+$ letters to the left of the sequence, so that it always runs the \emph{Move+} rule for ever. Otherwise, the sequence consumes $+$ letters until some of the following five situations take place: 

\begin{align*}
(a)~ \mathrm{if~}  w(k)_{j1: j1+l+l'+2}=~ & + [seq', l'] + [seq,l]+ \\
 \mathrm{then ~}  w(k+1)_{j1: j1+l+l'+2}=~ & [seq', l'] + [seq,l]+ (+)
\end{align*}
Both sequences \emph{Move+} to the left, until something happens to $[seq',l']$ that makes $[seq,l]$ act accordingly. 

When the orientations are balanced, $n_+ = n_-$, i.e., there are as many $+$ as $-$. Thus, at some point the sequence $[seq',l']$ will meet a $-$ and will act as described next in $(b)$. 

When the orientations are unbalanced, $n_+>n_-$, i.e., there are more $+$ letters than $-$ letters. Sequence $[seq',l']$ may meet a $-$, behaving both sequences as described next in $(b)$. It may be also the case that sequence $[seq',l']$ only meets $+$ letters; then, both sequences will behave for ever running the \emph{Move+} rule. 

\begin{align*}
(b)~ \mathrm{if~}  w(k)_{j1: j1+l+l'+2}=~ &- [seq', l'] + [seq,l]+   ~ \mathrm{then}\\
  w(k+1)_{j1: j1+l+l'+2}=~ & (-) -[seq', l'-2] + [seq,l]+ (+)
\end{align*}
Sequence $[seq',l']$ \emph{Reduces} (Lemma~\ref{lemma_collaborative_sequences_reduce}), feeding with additional $+$ to $[seq,l]$ until $[seq',l']$ disappears and $[seq,l]$ finds the first $-$. This case $(e)$ is explained later. 

\begin{align*}
(c)~ \mathrm{if~}  w(k)_{j1: j1+l+l'+3}=&&& + [seq', l']- + [seq,l]+ \\
\mathrm{then~}w(k+1)_{j1: j1+l+l'+3}=&&&  [~~seq',  ~l'+l+2]+ (+)
\end{align*}
Sequence $[seq',l']$ \emph{Expands}, $[seq,l]$ \emph{Move+}, and then both \emph{Merge} in a larger sequence. Note this merged sequence has a $+$ to the right. Thus (Lemma~\ref{lemma_sequence_evolution_rules}), it will  either \emph{Reduce} (Lemma~\ref{lemma_collaborative_sequences_reduce}) or start performing \emph{Move+} (behaving according to the different cases discussed in this proof). 

\begin{align*}
(d)~ \mathrm{if~}  w(k)_{j1: j1+l+l'+3}= &&& - [seq', l']- + [seq,l]+ \\
\mathrm{then~}w(k+1)_{j1: j1+l+l'+3}=&&&  (-) - [seq',  l' + l] + (+)
\end{align*}
 Sequence $[seq',l']$ \emph{Move-}, $[seq,l]$ \emph{Move+}, and then both \emph{Merge} in a larger sequence $[seq',  l' + l]$. Since this sequence has a letter $-$ to the left and $+$ to the right, then it will \emph{Reduce},  behaving as discussed in Lemma~\ref{lemma_collaborative_sequences_reduce}.

\begin{align*}
(e)~ \mathrm{if~}  w(k)_{j1: j1+l+3}= &&& -~-+[seq, l~]~+\\
\mathrm{then~} w(k+1)_{j1: j1+l+3}= &&& (-)-[seq, l]+(+)\\ 
\end{align*}
Sequence $[seq,l]$ has some $-$ letters to its left that do not belong to sequences  
(thus, there are at least two consecutive $-$ letters). It makes a \emph{Move+}, and finishes with a $-$ to the left and $+$ to the right. Then, it will \emph{Reduce}, behaving as discussed in Lemma~\ref{lemma_collaborative_sequences_reduce}.
\end{IEEEproof}

\section{Proof of Proposition \ref{prop_always_expand_move}}
\label{appendices_prop_always_expand_move}

\begin{IEEEproof}[Proof of Proposition \ref{prop_always_expand_move}]

Sequences evolve according to the rules \emph{Move+}, \emph{Move-}, \emph{Expand}, \emph{Reduce}, and \emph{Merge} in Lemma~\ref{lemma_sequence_evolution_rules}. As stated by Lemmas~\ref{lemma_collaborative_sequences_reduce}, \ref{lemma_collaborative_sequences_move_neg}, if at some point a sequence experiences the \emph{Reduce}, or \emph{Move-} rule, then it will eventually  disappear. Moreover, this includes the  merging  between sequences where one of them has experienced the \emph{Reduce}, or \emph{Move-} rule.  

Thus, in order for a sequence to survive, it must either \emph{Expand} at every round, or \emph{Move+} during all the rounds, or be the result of merging a \emph{Expand} and a  \emph{Move+} sequences. On the other hand, sequences that \emph{Move+} keep their length unchanged, and only sequences that always \emph{Expand} increase the length of the sequence (Lemma~\ref{lemma_sequence_evolution_rules}). 

As proved next, as long as there are $-$ letters not belonging to \emph{Move+} or to \emph{Expand} sequences, one \emph{Expand} sequence survives. Note from Lemmas~\ref{lemma_collaborative_sequences_reduce}, \ref{lemma_collaborative_sequences_move_neg} that sequences that \emph{Reduce}, or \emph{Move-} collaborate to partially sort the letters, placing the $-$ to the left and the $+$ to the right. Note also that if a \emph{Move+} sequence finds a $-$, it \emph{Reduces} (Lemma~\ref{lemma_collaborative_sequences_move_pos} and its proof).  

Thus, if all the \emph{Expand} sequences expire, when the last collaborative sequence (\emph{Reduce}, or \emph{Move-}) expires, it leaves all the letters sorted: $n_-$ $-$ letters followed by $n_+$ $+$ letters, e.g., 
 $+ + - - - -  + + $ (recall the cycle graph structure). Thus, a new sequence starting with $+-$ would appear (e.g., between positions 2 and 3 in the example). However, as stated by Lemma~\ref{lemma_sequence_properties}, no new sequences are created. Thus, a \emph{Expand} sequence must remain alive in order to consume these $+$ and $-$ letters. This concludes ($i$).

To prove ($ii$), it is noted that at least one sequence $[seq,l]$ \emph{Expands} at every round, increasing its length by 2 at every round, until its length together with the lengths of the \emph{Move+} sequences, sum up to $2 n_{bal}$. Thus, this takes $(2 n_{bal}-l)/2$ rounds, with $l\geq 2$, so that the number of rounds required to achieve the balanced configuration is smaller than $n_{bal}$.

The proof of ($iii$) is immediate from ($i$): the sequences lengths sum up to $2n_{bal}$ which equals $n$ in the balanced case. Thus, there are no letters in $w(k)$ and all the sequences must be together forming a single sequence.

The proof of ($iv$) follows from ($i$): since $w(k)$ only contains sequences and $+$ letters,  then sequences behave according to rule \emph{Move+} (Lemma~\ref{lemma_sequence_evolution_rules}).

Finally ($v$) follows from the previous properties and from the definition of interlaced orientations (Def.~\ref{def_interlaced}).
\end{IEEEproof}

\section{Proof of Theorems \ref{th_common_traversing_times}, \ref{th_performance_balanced} and \ref{th_performance_unbalanced}}
\label{appendices_ths}

\begin{IEEEproof}[Proof of Theorem ~\ref{th_common_traversing_times}]
It follows from Propositions \ref{prop_weightedConsensus} and \ref{prop_jointconn}.
\end{IEEEproof}

\begin{IEEEproof}[Proof of Theorem~\ref{th_performance_balanced}]

As stated by Proposition~\ref{prop_always_expand_move}, when robot regions have common  traversing times $t_\star$ and they have balanced orientations, then the robot orientations acquire an interlaced  configuration (Def. \ref{def_interlaced}) in less than  $n_{bal}=n/2$ rounds (Defs.  \ref{def_balanced_unbalanced}, \ref{def_round}). 
And from Proposition~\ref{prop_balanced_interlaced_clock_synch}, robots running the method  with balanced interlaced configurations and common traversing times associated to their  regions, synchronize in less than $n/2$ rounds (Def.\ref{def_round}) to a  configuration where they perform their meetings simultaneously in the network.
\end{IEEEproof}

\begin{IEEEproof}[Proof of Theorem~\ref{th_performance_unbalanced}]

As stated by Proposition~\ref{prop_always_expand_move}, when robot regions have common  traversing times $t_\star$, then the robot orientations acquire an interlaced  configuration (Def. \ref{def_interlaced}) in less than  $n_{bal}$ \emph{rounds} (Defs.  \ref{def_balanced_unbalanced}, \ref{def_round}). 
And from Proposition~\ref{prop_performance_unbalanced_interlaced}, a robot team running the method with unbalanced  interlaced configurations and common traversing times associated to their regions, 
has a performance in terms of the revisiting times $t_{rev}$ (Def.~\ref{def_revisiting_time}) averaged along $n_{bal}$ meetings given by ~\eqref{eq_th_perfomance_unbalanced}.
\end{IEEEproof}

\bibliographystyle{IEEEtran}

\end{document}